\pdfoutput=1

\documentclass[10pt,twocolumn,letterpaper]{article}

\usepackage[pagenumbers]{cvpr}
\definecolor{cvprblue}{rgb}{0.21,0.49,0.74}
\usepackage[pagebackref,breaklinks,colorlinks,allcolors=cvprblue]{hyperref}
\usepackage{amsmath,amssymb,amsfonts,amsthm,mathtools}
\usepackage{graphicx}
\usepackage{booktabs}
\usepackage{tikz,pgfplots,pgfplotstable}
\usetikzlibrary{positioning,calc}
\usepgfplotslibrary{groupplots}
\usepgfplotslibrary{fillbetween}
\usepackage{epstopdf}
\usepackage{placeins}
\usepackage{algorithm}
\usepackage[noend]{algpseudocode}
\algrenewcommand\algorithmicrequire{\textbf{Require:}}
\algrenewcommand\algorithmicindent{0.9em}
\definecolor{cbOrange}{RGB}{230,159,0}
\definecolor{cbSky}{RGB}{86,180,233}
\definecolor{cbGreen}{RGB}{0,158,115}
\definecolor{cbBlue}{RGB}{0,114,178}
\definecolor{cbVerm}{RGB}{213,94,0}
\definecolor{cbPurple}{RGB}{204,121,167}
\pgfplotsset{compat=1.17,
  mswbase/.style={height=3.6cm,
    log ticks with fixed point,
    tick label style={font=\tiny}, label style={font=\scriptsize},
    title style={font=\scriptsize, yshift=-2pt},
    every axis title/.append style={font=\scriptsize},
    xlabel style={yshift=3pt},
    axis line style={black!60}, tick style={black!60},
    grid=major, grid style={black!12, thin},
    every axis plot/.append style={line width=0.8pt, mark size=1.1pt},
    legend style={font=\tiny, /tikz/every even column/.append style={column sep=6pt}},
    clip mode=individual},
  msw/.style={mswbase, xmode=log, ymode=log, ymin=0.05, ymax=9000},
  cours/.style={black, line width=1.4pt, mark=*, mark size=1.3pt},
  cfid/.style={cbVerm, mark=square*},
  ckid/.style={cbGreen, mark=triangle*},
  ccmmd/.style={cbBlue, mark=diamond*},
  cc2st/.style={cbPurple, mark=o},
  csw/.style={black!45, densely dotted, mark=x},
  ccsw/.style={cbSky, densely dashed, mark=+},
  ccov/.style={cbOrange, mark=asterisk},
  cstd/.style={black, line width=1.1pt, densely dashed, mark=*, mark options={fill=white}},
  civ/.style={black!45, line width=0.8pt, mark=triangle, mark options={fill=white}},
  cpswd/.style={cbVerm!60!black, densely dashed, mark=pentagon, mark options={fill=white, solid}},
  cspec/.style={cbSky!70!black, densely dashdotted, mark=pentagon*, mark size=1.2pt},
  cps/.style={cbGreen!55!black, dash dot dot, mark=halfcircle*, mark size=1.2pt},
  cdino/.style={cbVerm, densely dashdotted, mark=square, mark options={fill=white, solid}},
  thr/.style={black!70, dashed, thin, no marks, forget plot},
  guide/.style={black!50, dash dot, thin, no marks, forget plot}}

\newcommand{\R}{\mathbb{R}}

\newcommand{\E}{\mathbb{E}}
\newcommand{\dd}{{\rm d}}
\newcommand{\abs}[1]{\left\lvert#1\right\rvert}
\newcommand{\norm}[1]{\left\lVert#1\right\rVert}
\def\V#1{{\boldsymbol{#1}}}
\def\M#1{{\bf{#1}}}
\newcommand{\thetab}{\boldsymbol{\theta}}

\renewcommand{\S}{\mathbb{S}}

\newtheorem{proposition}{Proposition}

 \newcommand{\colgroup}[1]{\textcolor{black!50}{#1}}

\title{MSW: a Multi-Scale Sliced Wasserstein Distance Between Image Distributions}

\author{Youssef Haouchat, Stanislas Ducotterd, Bassam El Rawas, Michael Unser\\
Biomedical Imaging Group, EPFL, Lausanne, Switzerland}

\begin{document}
\makeatletter\patchcmd{\@maketitle}{\vskip .375in}{\vskip 13.2pt}{}{\PATCHFAILED}\makeatother
\maketitle

\begin{abstract}
We propose the multi-scale sliced Wasserstein (MSW) distance between image distributions, which operates in pixel space and requires no trained network. MSW convolves images with a fixed bank of discrete cosine filters at several scales, pools the responses over all positions, and compares the resulting one-dimensional distributions using the Wasserstein distance. We pair MSW with an exact two-sample test that gives calibrated $p$-values. Most metrics compare two image sets in the latent space of a pretrained network. We show that these metrics are insensitive to simple perturbations of the images, because the latent space is nearly invariant to them. In contrast, MSW detects smaller perturbations than latent-space metrics and needs far fewer images to do so. Although it uses no learned features, it still detects semantic changes of the image distribution. Finally, MSW matches or outperforms existing Wasserstein-based metrics in our experiments.
\end{abstract}

\section{Introduction}
\label{sec:intro}
\begin{figure*}[t]
\centering
\resizebox{\textwidth}{!}{\begin{tikzpicture}
\begin{groupplot}[group style={group size=6 by 1, horizontal sep=3.5mm,
                  ylabels at=edge left, yticklabels at=edge left},
  mswbase, width=4cm, height=3.1cm, grid=none,
  ymin=0, enlarge y limits={upper, value=0.06}, axis x line*=bottom, axis line style={black!60}, ytick=\empty, ylabel style={align=center},
  unbounded coords=jump, no marks,
  title style={align=center, font=\scriptsize, yshift=5pt},
  legend style={draw=none, fill=none, font=\fontsize{5.5}{6}\selectfont, inner sep=1pt,
    at={(0.5,1.03)}, anchor=south,
    legend columns=3, /tikz/every even column/.append style={column sep=2.5pt},
    legend image code/.code={\draw[##1, line width=1pt] plot coordinates {(0cm,0cm)(0.22cm,0cm)};}},
  legend cell align=left]
\nextgroupplot[title={\strut spectral slope\\[-2pt]{\color{black!55}\strut textures}}, xmin=-3.2, xmax=3.2, xtick={-2,0,2}, axis y line*=left, ylabel={empirical\\[-2pt]marginal}, title style={name=firsttitle},]
\addplot[cbBlue!40, line width=0.6pt] table[x=x, y=c0, col sep=comma]{data/margrow_grf_slope.csv}; \addlegendentry{$0$}
\addplot[cbBlue!70, line width=0.6pt] table[x=x, y=c1, col sep=comma]{data/margrow_grf_slope.csv}; \addlegendentry{$0.26$}
\addplot[cbBlue!60!black, line width=0.6pt] table[x=x, y=c2, col sep=comma]{data/margrow_grf_slope.csv}; \addlegendentry{$1$}
\nextgroupplot[title={\strut spectral slope\\[-2pt]{\color{black!55}\strut heavy-tailed textures}}, xmin=-3.2, xmax=3.2, xtick={-2,0,2}, axis y line=none,]
\addplot[cbBlue!40, line width=0.6pt] table[x=x, y=c0, col sep=comma]{data/margrow_t3_slope.csv}; \addlegendentry{$0$}
\addplot[cbBlue!70, line width=0.6pt] table[x=x, y=c1, col sep=comma]{data/margrow_t3_slope.csv}; \addlegendentry{$0.26$}
\addplot[cbBlue!60!black, line width=0.6pt] table[x=x, y=c2, col sep=comma]{data/margrow_t3_slope.csv}; \addlegendentry{$1$}
\nextgroupplot[title={\strut heavier tails\\[-2pt]{\color{black!55}\strut textures}}, xmin=-4.2, xmax=4.2, xtick={-3,0,3}, axis y line=none,]
\addplot[cbBlue!40, line width=0.6pt] table[x=x, y=c0, col sep=comma]{data/margrow_grf_marginal.csv}; \addlegendentry{G}
\addplot[cbBlue!70, line width=0.6pt] table[x=x, y=c1, col sep=comma]{data/margrow_grf_marginal.csv}; \addlegendentry{$t_8$}
\addplot[cbBlue!60!black, line width=0.6pt] table[x=x, y=c2, col sep=comma]{data/margrow_grf_marginal.csv}; \addlegendentry{$t_3$}
\nextgroupplot[title={\strut noise\\[-2pt]{\color{black!55}\strut CelebA, red--green channel}}, xmin=-9, xmax=9, xtick={-6,0,6}, axis y line=none,]
\addplot[cbBlue!40, line width=0.6pt] table[x=x, y=c0, col sep=comma]{data/margrow_cel_noise.csv}; \addlegendentry{$0$}
\addplot[cbBlue!70, line width=0.6pt] table[x=x, y=c1, col sep=comma]{data/margrow_cel_noise.csv}; \addlegendentry{$0.0025$}
\addplot[cbBlue!60!black, line width=0.6pt] table[x=x, y=c2, col sep=comma]{data/margrow_cel_noise.csv}; \addlegendentry{$0.005$}
\nextgroupplot[title={\strut blur\\[-2pt]{\color{black!55}\strut CelebA}}, xmin=-1.6, xmax=1.6, xtick={-1,0,1}, axis y line=none,]
\addplot[cbBlue!40, line width=0.6pt] table[x=x, y=c0, col sep=comma]{data/margrow_cel_blur_050.csv}; \addlegendentry{$0$}
\addplot[cbBlue!70, line width=0.6pt] table[x=x, y=c1, col sep=comma]{data/margrow_cel_blur_050.csv}; \addlegendentry{$0.33$}
\addplot[cbBlue!60!black, line width=0.6pt] table[x=x, y=c2, col sep=comma]{data/margrow_cel_blur_050.csv}; \addlegendentry{$0.5$}
\nextgroupplot[title={\strut checkerboard\\[-2pt]{\color{black!55}\strut CIFAR-10}}, xmin=-9, xmax=9, xtick={-6,0,6}, axis y line=none,]
\addplot[cbBlue!40, line width=0.6pt] table[x=x, y=c0, col sep=comma]{data/margrow_cif_checker.csv}; \addlegendentry{$0$}
\addplot[cbBlue!70, line width=0.6pt] table[x=x, y=c1, col sep=comma]{data/margrow_cif_checker.csv}; \addlegendentry{$0.01$}
\addplot[cbBlue!60!black, line width=0.6pt] table[x=x, y=c2, col sep=comma]{data/margrow_cif_checker.csv}; \addlegendentry{$0.02$}
\end{groupplot}
\node[anchor=north east, align=right, font=\scriptsize\itshape, inner sep=0pt] at ($(firsttitle.north -| group c1r1.west) + (-1.5mm, -0.3333em)$)
  {\strut perturbation\\[-2pt]{\color{black!55}\strut data}};
\end{tikzpicture}
 }
\caption{Histograms of the responses of one high-frequency filter under six
perturbations (darker is stronger). The textures are synthetic
(Section~\ref{sec:more}).}
\label{fig:marginalsrow}
\end{figure*}
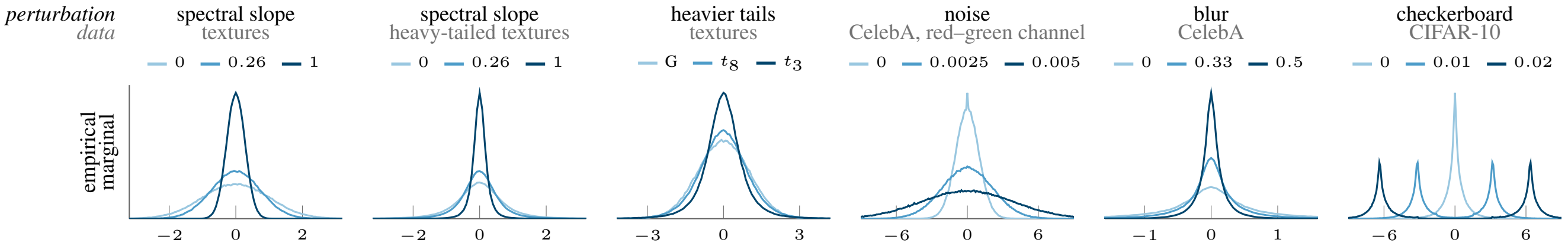
 Many tasks require a comparison of two sets of images. A generative
model, for instance, is evaluated by a distance between the distribution of
its outputs and that of real images. For images with $d$ pixel values, the
Wasserstein distance on $\R^d$ is expensive to compute, and its estimation
requires a number of images exponential in
$d$~\cite{peyreComputationalOptimalTransport2020}.
In practice, the images are therefore first projected onto a simpler space,
where distributions are easy to compare.

Latent-based metrics such as FID~\cite{heusel2017gans} and
CMMD~\cite{jayasumana2024rethinking} are the standard choice. Each image is mapped to a latent space by a pretrained
network, and the two sets of latent vectors are then compared using a distance, such as the Fr\'echet distance or a
maximum mean discrepancy. Such metrics are only as sensitive as their latent
spaces. The networks underlying FID and CMMD are trained to classify images
or to match them with text. They have no reason to keep the details of an
image that do not help with these tasks. Thus, their latent spaces are
nearly invariant to small perturbations such as a faint noise, a slight
blur or a faint periodic pattern. A latent-based metric detects
such a perturbation only when it is strong or when many images are
available, and it cannot detect a perturbation to which the latent space is
invariant.

We propose the multi-scale sliced Wasserstein (MSW) distance, which is
computed directly in pixel space and contains no trained component. Instead
of a latent space, MSW projects the images onto many one-dimensional spaces. It
convolves every image with a fixed bank of discrete cosine filters at different scales. A filter returns one value at every position of an image. For each
filter, the values from all positions and all images of a set form a single
histogram (Figure~\ref{fig:marginalsrow}). The histograms of the two sets are then
compared using the Wasserstein distance, which in one dimension reduces to
a sorting operation. The use of
every position is based on the approximate translation invariance of image
statistics~\cite{simoncelli2001natural}. It allows MSW to operate with few images since each image
contributes as many samples as it has positions instead of a single latent
vector.

We compare MSW with FID, CMMD and the convolution sliced Wasserstein distance
(CSW)~\cite{nguyen2022revisiting}. For most low-level
perturbations, such as noise, blur and periodic patterns,
MSW detects much smaller amplitudes than the latent-based metrics, and it
needs far fewer images. On Stable Diffusion images of increasing quality, it
recovers the order of the sets from a few images more reliably than they do.
On super-resolution networks, it agrees with them on the gain in realism
brought by adversarial training. Semantic changes are the hardest case for a
metric without learned features. MSW still separates nearly all pairs of
COCO classes. Across all these experiments, MSW also outperforms CSW, which
its authors report to improve on the conventional sliced Wasserstein
distance on images.
The code and the experiment scripts will be released upon acceptance.

\section{Related work}
\label{sec:related}
We consider two families of metrics. Latent-based metrics compare two image
sets in a latent space, and sliced Wasserstein distances compare
one-dimensional projections of two image sets directly in pixel space.

\textbf{Latent-based metrics.} FID~\cite{heusel2017gans}, the standard
metric for generated images, fits a Gaussian
to the Inception latent vectors of each set and reports the
$2$-Wasserstein distance between the two Gaussians.
CMMD~\cite{jayasumana2024rethinking}, designed to correct the weaknesses of
FID, computes a Gaussian-kernel maximum mean
discrepancy between CLIP latent vectors.

\textbf{Sliced Wasserstein distances.} The sliced Wasserstein
distance (SW)~\cite{rabin2011wasserstein,bonneel2015sliced} averages the Wasserstein distance between the projections of the two
distributions over uniformly random
directions. Applied to vectorized
images, random directions ignore the spatial structure of the data.
CSW~\cite{nguyen2022revisiting} accounts for this structure by projecting
each image with a sequence of random convolutions. We take CSW as the
baseline of this family, since it is designed for images.
Sliced distances between distributions of patches or network features were used earlier to evaluate
GANs~\cite{karras2018progressive} and as losses
for texture synthesis~\cite{heitz2021sliced,elnekave2022generating}.
He et al.~\cite{he2024multiscale} measure colour differences between two
images this way.
In our experiments, CSW and the patch distance of Karras et
al.~\cite{karras2018progressive} give similar results. MSW
uses a fixed and interpretable filter bank rather than random directions.
Unlike CSW, it keeps every position of a convolution output.
Such filter banks come from
texture analysis, where the histograms of their outputs describe the
local statistics of an image~\cite{unser1986local}.
The Fr\'echet wavelet distance~\cite{veeramacheneni2025frechet} also uses
fixed filters, but fits a Gaussian to their outputs.

\section{The MSW distance}
\label{sec:method}

\subsection{Background}
\label{sec:background}
For two probability distributions $\mu,\nu$ on $\R^d$ and for $p \geq 1$, the $p$-Wasserstein
distance is
\begin{equation}
\label{eq:wp}
W_p^p(\mu,\nu)=\min_{\gamma\in\Pi(\mu,\nu)}\int\norm{\V x-\V y}^p\,\dd\gamma(\V x,\V y),
\end{equation}
where $\Pi(\mu,\nu)$ is the set of couplings of $\mu$ and $\nu$, that is,
the joint distributions $\gamma$ on $\R^d\times\R^d$ whose marginals are
$\mu$ and $\nu$. In one dimension, $W_p$ reduces to a comparison of the quantile
functions\footnote{The quantile function $Q_\mu : [0,1] \to \R$ is defined in terms of the c.d.f. $F_\mu$ as $Q_\mu(t) = \inf\{x \in \R \,:\, F_\mu(x) \geq t\}$. When $F_\mu$ is strictly increasing and continuous, we have that $Q_\mu(t) = F_\mu^{-1}(t)$.} $Q_\mu$ and
$Q_\nu$~\cite[Prop.~2.17]{santambrogioOptimalTransportApplied2015},
\begin{equation}
\label{eq:w1d}
W_p^p(\mu,\nu)=\int_0^1\abs{Q_\mu(t)-Q_\nu(t)}^p\,\dd t,
\end{equation}
and for two empirical distributions with $n$ samples each, it equals the mean
of $\abs{x_{(i)}-y_{(i)}}^p$, where $x_{(i)}$ and $y_{(i)}$ are the $i$th
smallest samples. In higher dimension, the optimal transport problem in~\eqref{eq:wp} is
costly to solve. The distance between the empirical distributions
of $n$ samples also converges slowly to $W_p(\mu,\nu)$, at the rate
$\mathcal{O}(n^{-1/d})$~\cite{fournierRateConvergenceWasserstein2015}.

The sliced Wasserstein distance~\cite{rabin2011wasserstein,bonneel2015sliced}
avoids both issues. A \emph{direction} is a unit vector $\thetab\in\R^d$.
Let $P_{\thetab}(\V x)=\thetab^\top\V x$ denote the projection onto
$\thetab$, and $(P_{\thetab})_\sharp\mu$ the
distribution of $P_{\thetab}(\V x)$ for $\V x\sim\mu$. Then,
\begin{equation}
\label{eq:sw}
\mathrm{SW}_p^p(\mu,\nu)=\E_{\thetab}\Big[W_p^p\big((P_{\thetab})_\sharp\mu,(P_{\thetab})_\sharp\nu\big)\Big],
\end{equation}
where $\thetab$ is uniform on the unit sphere $\S^{d-1}$. By the
Cram\'er--Wold theorem~\cite{ref:cramerwold}, $\mathrm{SW}_p$ is a metric. Each
projection costs a sort, and a Monte-Carlo estimate with $J$ directions
converges at the rate $\mathcal{O}(J^{-1/2})$, independently of $d$. The
price is that a random direction mixes all coordinates and dilutes any
difference between $\mu$ and $\nu$. For $p=2$, it is known~\cite[Sec.~2.2]{park2025geometry} that
$\mathrm{SW}_2\le d^{-1/2}\,W_2$. For
images, the directions along which two distributions differ are far from
random.

\subsection{Choice of the directions and scales}
\label{sec:directions}
\begin{figure*}[t]
\centering
\usetikzlibrary{arrows.meta,decorations.pathreplacing}\colorlet{setA}{cbBlue}\colorlet{setB}{cbVerm}\newcommand{\mstack}[4]{\foreach \j/\o in {2/3, 1/1.5}{
    \fill[white] (#1+\o-0.35, #2+\o-0.35) rectangle ++(16.1, 16.1);
    \node[anchor=south west] at (#1+\o, #2+\o) {\includegraphics[width=15.4mm, height=15.4mm]{#3\j}};
    \draw[#4, line width=0.6pt] (#1+\o, #2+\o) rectangle ++(15.4, 15.4);}
  \fill[white] (#1-0.35, #2-0.35) rectangle ++(16.1, 16.1);
  \node[anchor=south west] at (#1, #2) {\includegraphics[width=15.4mm, height=15.4mm]{#30}};
  \draw[#4, line width=0.6pt] (#1, #2) rectangle ++(15.4, 15.4);}\begin{tikzpicture}[x=1mm, y=1mm, every node/.style={inner sep=0, outer sep=0},
    flow/.style={-{Stealth[length=1.7mm, width=1.5mm]}, black!40, line width=0.55pt},
    head/.style={font=\footnotesize, text=black!75, anchor=base},
    lab/.style={font=\scriptsize, text=black!60},
    tiny/.style={font=\fontsize{6.5}{7}\selectfont, text=black!60},
    sel/.style={black!75, line width=0.45pt}, unsel/.style={black!25, line width=0.3pt}]
\node[head] at (13.4, 48.4) {two image sets};
  \node[head] at (60.5, 48.4) {one slice $q=(s,\V h,\V c)$};
  \node[head] at (106.8, 48.4) {response maps $\Phi_q(\V x)$};
  \node[head] at (150.3, 48.4) {histograms over all images and positions};
\mstack{4.2}{25.8}{figures/method_A}{setA}
  \mstack{4.2}{3.2}{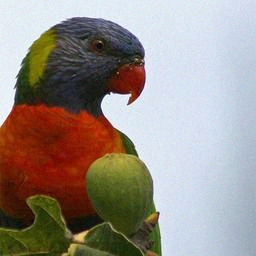}{setB}
  \node[setA, font=\small] at (1.5, 33.5) {$A$};
  \node[setB, font=\small] at (1.5, 10.9) {$B$};
  \node[tiny, text=setA, anchor=base] at (11.9, 23.1) {clean};
  \node[tiny, text=setB, anchor=base] at (11.9, 0.3) {slightly noisy};
  \draw[flow] (24.4, 33.5) -- (29.8, 33.5);
  \draw[flow] (24.4, 10.9) -- (29.8, 10.9);
\fill[black!5, rounded corners=1.5mm] (31, -1.5) rectangle (90, 45);
  \node[lab, anchor=base] at (45, 41) {scale $s$};
  \node[lab, anchor=base] at (63.5, 41) {filter $\V h$};
  \node[lab, anchor=base] at (81, 41) {channel $\V c$};
\node[anchor=south west] at (38.5, 22.4) {\includegraphics[width=13mm]{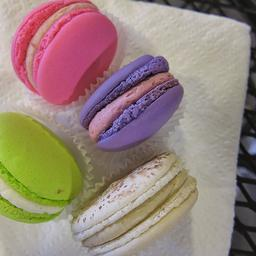}};
  \node[anchor=south west] at (41.75, 12.9) {\includegraphics[width=6.5mm]{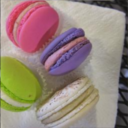}};
  \node[anchor=south west] at (43.375, 6.65) {\includegraphics[width=3.25mm]{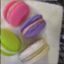}};
  \draw[sel] (38.5, 22.4) rectangle ++(13, 13);
  \draw[unsel] (41.75, 12.9) rectangle ++(6.5, 6.5);
  \draw[unsel] (43.375, 6.65) rectangle ++(3.25, 3.25);
  \node[lab, anchor=east] at (37.3, 28.9) {$0$};
  \node[lab, anchor=east] at (37.3, 16.15) {$1$};
  \node[lab, anchor=east] at (37.3, 8.3) {$2$};
\node[anchor=south west] at (55.5, 13.5) {\includegraphics[width=16mm]{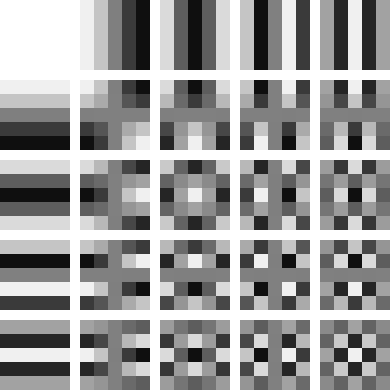}};
  \draw[sel] (55.5+1*3.282-0.2, 29.5-2.872-1*3.282-0.2) rectangle ++(3.272, 3.272);
  \node[lab, anchor=base] at (63.5, 9.2) {$5{\times}5$ DCT};
\foreach \f/\t/\y in {lum/luminance/29.9, rg/red--green/17.4, by/blue--yellow/4.9}{
    \node[anchor=south west] at (76.75, \y) {\includegraphics[width=8.5mm]{figures/method_c_\f}};
    \node[tiny, anchor=base] at (81, \y-2.55) {\t};}
  \draw[sel] (76.75, 29.9) rectangle ++(8.5, 8.5);
  \draw[unsel] (76.75, 17.4) rectangle ++(8.5, 8.5);
  \draw[unsel] (76.75, 4.9) rectangle ++(8.5, 8.5);
  \draw[flow] (91.2, 33.5) -- (96.6, 33.5);
  \draw[flow] (91.2, 10.9) -- (96.6, 10.9);
\mstack{98}{25.8}{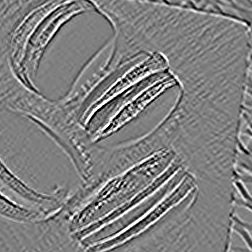}{setA}
  \mstack{98}{3.2}{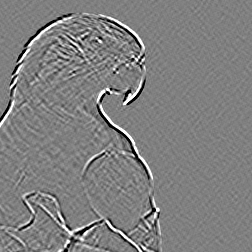}{setB}
\draw[setA, line width=0.6pt, decorate, decoration={brace, amplitude=1.2mm}] (117.9, 44.5) -- (117.9, 25.5);
  \draw[setB, line width=0.6pt, decorate, decoration={brace, amplitude=1.2mm}] (117.9, 21.9) -- (117.9, 2.9);
  \draw[flow] (119.9, 35) to[out=0, in=180] (128.2, 28.5);
  \draw[flow] (119.9, 12.4) to[out=0, in=180] (128.2, 23);
\foreach \o in {2.6, 1.3}
    \draw[black!30, line width=0.4pt, fill=white, rounded corners=0.8mm] (129+\o, 12.5+\o) rectangle ++(42.6, 27);
  \draw[black!40, line width=0.4pt, fill=white, rounded corners=0.8mm] (129, 12.5) rectangle ++(42.6, 27);
  \begin{axis}[at={(131.6mm,17.4mm)}, anchor=south west, scale only axis, width=37.4mm, height=19mm,
      axis x line=bottom, axis y line=none, axis line style={black!55, -}, tick style={black!55},
      xmin=-0.123, xmax=0.123, ymin=0, ymax=1.06, xtick={0}, xticklabels={}, major tick length=1mm, scaled ticks=false,
      xlabel={response value},
      label style={font=\scriptsize}, xlabel style={yshift=-1.5pt}, clip=false, no marks]
    \addplot[draw=none, fill=setB, fill opacity=0.13] table[x=x, y=b, col sep=comma]{data/method_hist.csv} \closedcycle;
    \addplot[draw=none, fill=setA, fill opacity=0.13] table[x=x, y=a, col sep=comma]{data/method_hist.csv} \closedcycle;
    \addplot[setB, line width=0.9pt] table[x=x, y=b, col sep=comma]{data/method_hist.csv};
    \addplot[setA, line width=0.9pt] table[x=x, y=a, col sep=comma]{data/method_hist.csv};
    \node[setA, font=\small, anchor=west] at (axis cs:0.008, 0.86) {$\mu_q$};
    \node[setB, font=\small, anchor=south west] at (axis cs:0.021, 0.36) {$\nu_q$};
  \end{axis}
\draw[flow] (150.3, 11.7) -- (150.3, 7.6);
  \node[lab, anchor=west] at (151.8, 9.7) {all slices $q$};
  \node[font=\scriptsize, anchor=base] at (150.3, 1.8)
    {$\mathrm{MSW}(\mu,\nu)=\big(\sum_q\lambda_q\,W_2^2(\textcolor{setA}{\mu_q},\textcolor{setB}{\nu_q})\big)^{1/2}$};
\end{tikzpicture}
 \caption{Overview of MSW: for each slice, the responses of all images and positions form one histogram per set.}
\label{fig:method}
\vspace{12pt}
\end{figure*}
 When $\mu$ has a density, the one-dimensional densities of its projections
$(P_{\thetab})_\sharp\mu$ over all directions $\thetab$ form the Radon
transform of $\mu$~\cite{bonneel2015sliced,kolouri2019generalized}. As in
tomography, each projection is a view of $\mu$ from one angle. The sliced
Wasserstein distance compares $\mu$ and $\nu$ one projection at a time.
The full set determines $\mu$, but only finitely many can be computed.

If the projections of an image onto two directions are correlated, the
second direction only tells little more than the first. We therefore want
directions with uncorrelated projections. The pixels themselves are a poor
choice. Neighbouring pixels of natural images are indeed highly correlated, and this correlation
extends over long
distances~\cite{field1987relations,simoncelli2001natural}. The statistics
of natural images are also approximately invariant to translation. For a
small patch, the basis of the discrete cosine transform (DCT) then gives
nearly uncorrelated
projections~\cite{ahmed1974discrete,unser1984approximation}. This basis is
the same for every set of natural images, and MSW takes its directions from
it.

A small patch misses the long-range correlation. Natural images have
nearly the same statistics at every
scale~\cite{field1987relations,ruderman1994statistics}. We
therefore apply the same filters to coarser versions of the
image, as in a multiresolution decomposition~\cite{mallat1989theory}. At a
coarse scale, a filter covers a large part of the image.

\subsection{Multiscale sliced Wasserstein distance}
\label{sec:definition}
Let $\V x\in\R^{C\times H\times W}$ be an image with $C$ channels, and let
$\V x_s$ denote the image at scale $s>0$, obtained from $\V x$ by $s$
successive blur-and-decimate steps, with $\V x_0=\V x$. A \emph{slice}
$q=(s,\V h,\V c)$ consists of a scale $s$, a spatial filter
$\V h\in\R^{K\times K}$ and a channel vector $\V c\in\R^C$. It defines the
linear map
\begin{equation}
\label{eq:slice}
\Phi_q(\V x)=\V h*\big(\V c^\top\V x_s\big),
\end{equation}
where $\V c^\top$ combines the channels and $*$ is the two-dimensional
convolution. The
\emph{response map} $\Phi_q(\V x)$ is an image with $n_s$ pixels, and
$\Phi_q(\V x)[\V k]$ is its value at the pixel $\V k$, which we call a
\emph{position}. Each value is a linear functional of $\V x$, and we
normalize $\V h$ so that this functional has unit norm.
Section~\ref{sec:bank} specifies the filters and the channel vectors.

For a distribution $\mu$ over images, the \emph{slice distribution} $\mu_q$
collects the responses of slice $q$ at all positions. Formally, $\mu_q$ is
the average, over the $n_s$ positions $\V k$, of the distribution of
$\Phi_q(\V x)[\V k]$ for $\V x\sim\mu$. A set of $N$ images thus provides
$N n_s$ samples of it.
Figure~\ref{fig:marginalsrow} shows the histograms of such samples. The MSW
distance between two distributions $\mu$ and $\nu$ over images is
\begin{equation}
\label{eq:msw}
\mathrm{MSW}(\mu,\nu)=\Big(\sum_q\lambda_q\,W_2^2\big(\mu_q,\nu_q\big)\Big)^{1/2},
\end{equation}
with weights $\lambda_q\propto4^{-s}/\sigma_q$ normalized to sum to
one. The factor $4^{-s}$ is the fraction of positions contributed by scale
$s$, and $\sigma_q$ is the standard deviation of the slice distribution of
a fixed reference set of natural images, measured once. This normalization
brings all slices to a comparable scale without making the weights depend on
$\mu$ or $\nu$. Figure~\ref{fig:method} summarizes the construction.
Each term of~\eqref{eq:msw} belongs to one slice. The terms can be read one
by one, and they show at which scale, in which frequency band and along
which colour axis two sets of images differ.

MSW is a sliced Wasserstein distance with fixed directions, applied to
patches. The samples are the $K\times K$ patches of $\V x_s$ at all
positions, and the filters and the channel vectors give the directions.
A patch has only $CK^2$ dimensions, so that a complete basis of directions
is affordable. This is not the case for whole images.
When $\mu$ is stationary, $\mu_q$ is the distribution of the response at
any position, and pooling over positions is most effective. Otherwise, without assuming stationarity,
MSW still applies, and $\mu_q$ is
the average of the distributions at the $n_s$ positions. This average
describes the local statistics of $\mu$, but not where they occur. Such statistics remain informative on photographs,
which are not stationary~\cite{ref:torralba2003,ref:bagnet}. The use of
all positions makes MSW usable with few images. The number of samples per slice grows with the number of
positions and with the number of images, whereas a latent-based
metric obtains one vector per image and CSW one scalar per image and
projection.

\begin{table*}[t]
\centering\small
\caption{Ten controlled perturbations, with detection as defined in~\eqref{eq:power}, by the exact test for MSW (--: no detection).}
\label{tab:more}
\setlength{\tabcolsep}{4.5pt}
\begin{tabular*}{\textwidth}{@{\extracolsep{\fill}}llcccc@{\hspace{3.2em}}ccccc@{}}
\toprule
 & & \multicolumn{4}{c}{detection threshold $\varepsilon_{0.8}$ ($N{=}500$)} & \multicolumn{5}{c}{sample requirement $N^\ast$ at amplitude $\varepsilon$} \\
\cmidrule(r{2.4em}){3-6}\cmidrule{7-11}
 & & FID & CMMD & CSW & MSW & $\varepsilon$ & FID & CMMD & CSW & MSW \\
\midrule
CelebA & Gaussian noise & 0.01 & 0.0025 & -- & \textbf{0.0005} & 0.0025 & -- & 128 & -- & \textbf{8} \\
 & Gaussian blur & 0.33 & 0.22 & -- & \textbf{0.10} & 0.33 & 256 & 128 & -- & \textbf{8} \\
\addlinespace
COCO & grey checkerboard & 0.04 & 0.04 & 0.04 & \textbf{0.005} & 0.04 & 512 & 256 & 256 & \textbf{8} \\
 & red--green checkerboard & 0.16 & 0.08 & 0.08 & \textbf{0.01} & 0.08 & -- & 256 & 512 & \textbf{8} \\
 & shading & 0.16 & \textbf{0.04} & 0.16 & 0.08 & 0.08 & 1024 & \textbf{128} & -- & 256 \\
\addlinespace
CIFAR-10 & Gaussian blur & 0.5 & 0.22 & -- & \textbf{0.15} & 0.33 & 512 & 64 & -- & \textbf{16} \\
\addlinespace
synthetic textures & spectral slope & 0.128 & 0.064 & -- & \textbf{0.008} & 0.064 & 1024 & 512 & -- & \textbf{8} \\
 & white noise & 0.16 & 0.08 & -- & \textbf{0.04} & 0.08 & -- & 256 & -- & \textbf{8} \\
 & heavier tails & 1/12 & 1/12 & 1/12 & \textbf{1/48} & 1/12 & 512 & 512 & 512 & \textbf{128} \\
\addlinespace
heavy-tailed textures & spectral slope & 0.128 & 0.128 & -- & \textbf{0.032} & 0.128 & 128 & 128 & -- & \textbf{64} \\
\bottomrule
\end{tabular*}
\end{table*}

MSW is a pseudometric. It is symmetric, vanishes for $\mu=\nu$ and satisfies
the triangle inequality, since each term of~\eqref{eq:msw} is a Wasserstein
distance between two distributions obtained from $\mu$ and $\nu$ by the same
map. It is not a
metric, because the bank has finitely many filters and two distributions with
identical slice distributions are not distinguished. Since each slice map is
a mixture of unit-norm linear functionals and the weights sum to one,
$\mathrm{MSW}(\mu,\nu)\le W_2(\mu,\nu)$ (Appendix~\ref{app:estimation}).

In practice, $\mu$ and $\nu$ are known only through two
sets $A$ and $B$ of $N$ images each, and $\mu_q$ and $\nu_q$ are replaced by
the empirical distributions of the responses. Computed between two finite
samples, $W_2^2$ is biased upward. The same bias makes
FID~\cite{chong2020effectively} and CSW depend on the number of
images. Like CMMD, we use an estimator whose
expectation is exactly zero when $\mu=\nu$. The detection rule of
Section~\ref{sec:experiments} is an exact test of the hypothesis
$\mu=\nu$. The distance~\eqref{eq:msw} uses fixed weights, and this test uses
data-dependent
weights~\cite{cochran1954combination,biggs2023mmdfuse}
(Appendix~\ref{app:estimation}).

\subsection{The filter bank}
\label{sec:bank}
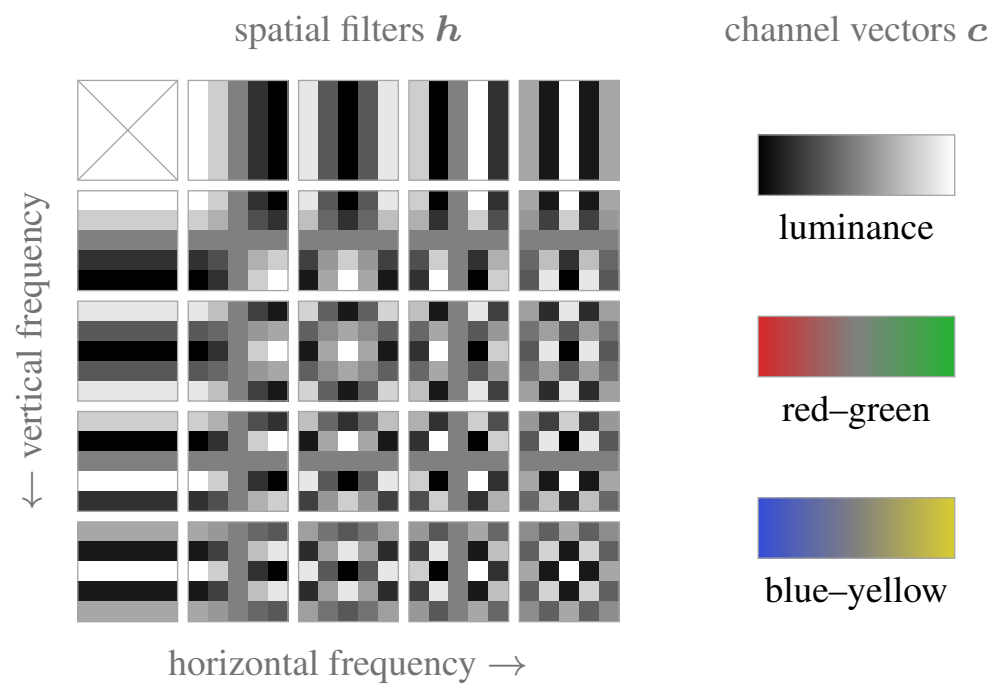
\begin{figure}[t]
\centering
\begin{tikzpicture}[x=1mm, y=1mm, every node/.style={inner sep=0},
    head/.style={font=\scriptsize, text=black!55, anchor=base}, name/.style={font=\scriptsize, anchor=base}]
  \def\atomside{6.6}\def\atompitch{7.3}
\foreach \a/\ma in {0/1, 1/0.951057, 2/1, 3/0.951057, 4/1}
    \foreach \c/\mc in {0/1, 1/0.951057, 2/1, 3/0.951057, 4/1} {
      \ifnum\numexpr\a+\c>0
        \foreach \i in {0,...,4} \foreach \j in {0,...,4} {
          \pgfmathsetmacro\g{min(100, max(0, 50-50*cos(18*(2*\i+1)*\a)*cos(18*(2*\j+1)*\c)/(\ma*\mc)))}
          \fill[black!\g] (\atompitch*\c+\atomside*\j/5, -\atompitch*\a-\atomside*\i/5)
            rectangle ++(\atomside/5+0.02, -\atomside/5-0.02);}
      \fi
      \draw[black!35, line width=0.25pt] (\atompitch*\c, -\atompitch*\a) rectangle ++(\atomside, -\atomside);}
\draw[black!35, line width=0.25pt] (0, 0) -- (\atomside, -\atomside) (0, -\atomside) -- (\atomside, 0);
  \node[head] at (17.9, 2.6) {spatial filters $\V h$};
  \node[font=\scriptsize, text=black!55, anchor=north] at (17.9, -37.562) {horizontal frequency $\rightarrow$};
  \node[font=\scriptsize, text=black!55, rotate=90, anchor=south] at (-1.8, -17.9) {$\leftarrow$ vertical frequency};
\node[head] at (51.5, 2.6) {channel vectors $\V c$};
  \definecolor{axR}{HTML}{D82626}\definecolor{axG}{HTML}{26B233}
  \definecolor{axB}{HTML}{334CD8}\definecolor{axY}{HTML}{D8CC33}
  \foreach \y/\l/\r/\t in {-3.6/black/white/luminance, -15.6/axR/axG/red--green, -27.6/axB/axY/blue--yellow} {
    \shade[left color=\l, right color=\r, middle color=black!50] (45, \y) rectangle ++(13, -4);
    \draw[black!35, line width=0.25pt] (45, \y) rectangle ++(13, -4);
    \node[name] at (51.5, \y-6.9) {\t};}
\end{tikzpicture}
 \caption{The $5{\times}5$ DCT atoms and the three channel axes.}
\label{fig:atlas}
\end{figure}
 A slice combines a filter, a channel vector and a scale.

\textbf{Spatial filters.} The filters $\V h$ are the $24$ non-constant atoms of
the $5\times5$ DCT (Figure~\ref{fig:atlas}).
Computed in a sliding window, the DCT is a bank of convolution
filters~\cite{unser1986local}, and its responses at a pixel are nearly
uncorrelated (Section~\ref{sec:directions}).
The atoms also tile the frequency plane, so that each slice has a scale and
a frequency band.

\textbf{Channel vectors.} The vectors $\V c$ are the three opponent colour axes of Figure~\ref{fig:atlas},
luminance $(1,1,1)/\sqrt3$, red--green $(1,-1,0)/\sqrt2$ and blue--yellow
$(1,1,-2)/\sqrt6$, which approximately decorrelate the colour channels of
natural images~\cite{ruderman1998statistics}.
The bank therefore comprises $72$ slices per scale. With the three constant
patches, their directions form an orthogonal basis of the $5\times5$ colour
patches. Natural images carry little variance on the
chromatic axes, where small perturbations stand out.

\textbf{Scales.} From one scale to the next, the image is smoothed with the
binomial filter $(1,4,6,4,1)/16$~\cite{unser1999splines} along each axis
and downsampled by two. All scales
larger than the filters are used, three at $32$ pixels and six
at $256$.

\section{Experiments}
\label{sec:experiments}
\label{sec:results}
\label{sec:protocol}
\label{sec:small}
The experiments go from low-level changes of the images to semantic ones.
Sections~\ref{sec:controlled} and~\ref{sec:detail} study controlled
perturbations, Sections~\ref{sec:sd} and~\ref{sec:restoration} compare
generators and networks, and Section~\ref{sec:semantic} compares classes.

\subsection{Protocol}
\label{sec:setup}
All experiments share the same metrics and detection rule.

\textbf{Metrics.} We compare MSW with FID, CMMD and CSW.
FID and CMMD use the publicly released Inception and CLIP networks
(Appendix~\ref{app:protocol}).

\textbf{Detection rule.} Let $D$ be any of the four metrics. One run draws
two independent sets $A$ and $A_\varepsilon$ of $N$ images at random from
the dataset, and perturbs the images of $A_\varepsilon$ with amplitude
$\varepsilon$. Since $D(A,A_0)$ varies between draws, the
perturbation is considered detected when $D(A,A_\varepsilon)$ exceeds $t_D$, the
$95$th percentile of $D(A,A_0)$. The
\emph{power} of $D$ is its probability of detection over independent runs,
that is, the share of runs with a $p$-value below $0.05$,
\begin{equation}
\label{eq:power}
\mathrm{power}_D(\varepsilon,N)=\mathbb{P}\big[D(A,A_\varepsilon)>t_D\big].
\end{equation}
For FID, CMMD and CSW, we estimate $t_D$ from unperturbed runs and count the
perturbed runs that exceed it (Appendix~\ref{app:protocol}). For MSW, the
exact test of Appendix~\ref{app:estimation} gives the $p$-value directly,
without unperturbed sets. This test weights the slices of~\eqref{eq:msw}
according to their variance.

Two quantities summarize each experiment. The \emph{detection threshold}
$\varepsilon_{0.8}$ is the smallest amplitude $\varepsilon$ such
that $\mathrm{power}_D(\varepsilon,500)\ge0.8$. The \emph{sample
requirement} $N^\ast$ is, at a fixed amplitude $\varepsilon$, the smallest
$N$ such that $\mathrm{power}_D(\varepsilon,N')\ge0.8$ for all
$N'\ge N$. The level $0.05$ and the power $0.8$ are
conventional~\cite{cohen1988statistical}.

\textbf{Cost.} MSW is more than ten times faster to evaluate than FID or
CMMD, and several times faster when its test is included. CSW is the fastest
(Appendix~\ref{app:protocol}).

\subsection{Controlled perturbations}
\label{sec:controlled}
\label{sec:noise}
\label{sec:checker}
\label{sec:more}
\begin{figure}[t]
\centering
\begin{tikzpicture}[x=1mm, y=1mm]
  \foreach \k/\f/\t in {0/image/image, 1/clean/detail, 2/pert/with the pattern,
                        3/diff/pattern {\color{black!55}($\times16$)}} {
    \node[inner sep=0] at (19.37*\k, 0) {\includegraphics[width=17.3mm, height=17.3mm]{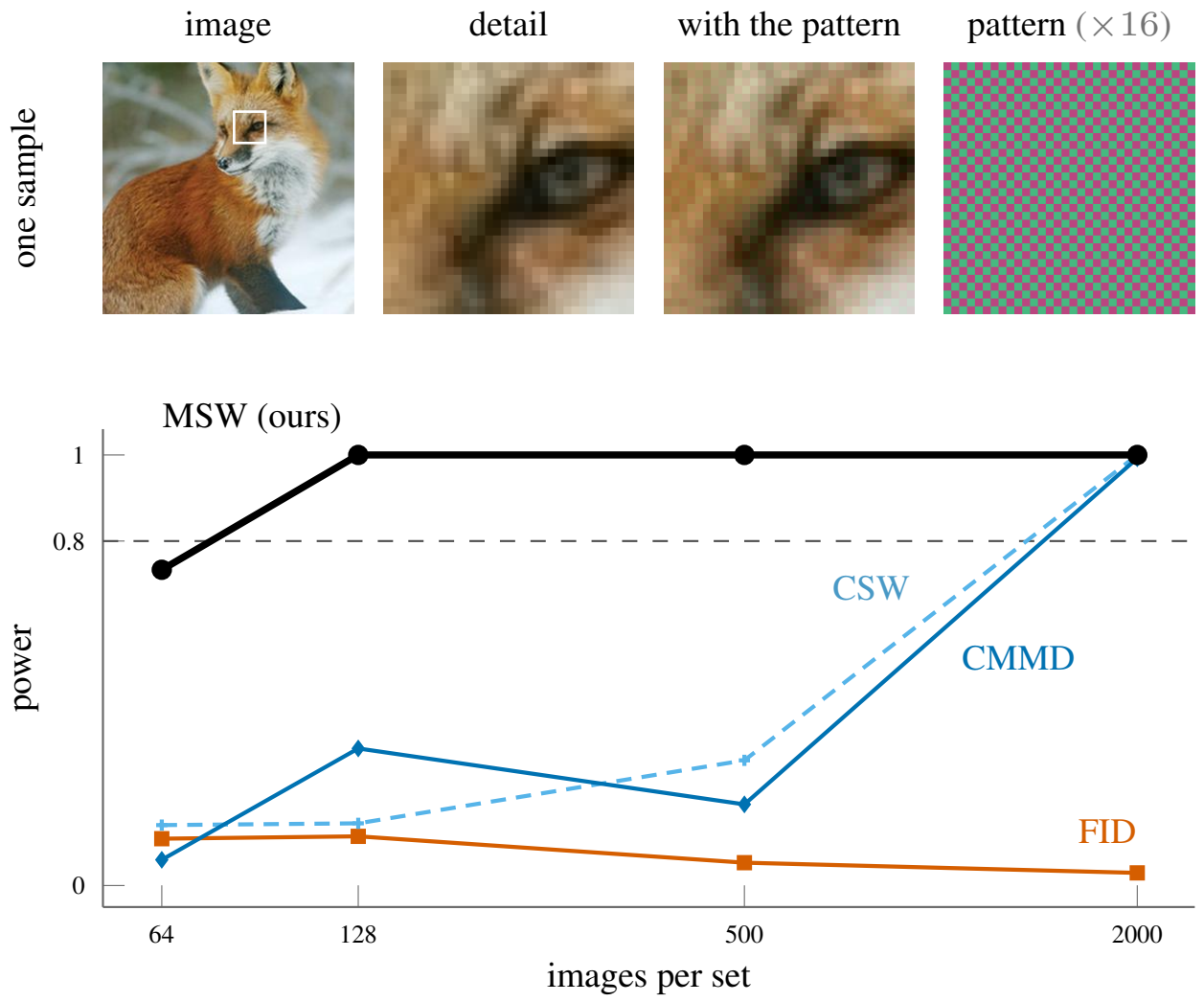}};
    \node[inner sep=0, font=\scriptsize, anchor=base] at (19.37*\k, 10.55) {\strut\t};
  }
  \draw[white, line width=0.5pt] (0.405, 5.271) rectangle (2.568, 3.109);
  \begin{axis}[mswbase, grid=none, at={(-8.65mm, -16.65mm)}, anchor=north west,
    scale only axis, width=75.4mm, height=33mm,
    axis x line*=bottom, axis y line*=left,
    xmode=log, xmin=52, xmax=2450, ymin=-0.05, ymax=1.06,
    xtick={64,128,500,2000}, xticklabels={64,128,500,2000}, minor tick style={draw=none},
    ytick={0,0.8,1}, yticklabels={0,0.8,1},
    xlabel={images per set}, ylabel={power}, ylabel style={yshift=-4pt, name=ylab},
    clip=false]
  \addplot[thr] coordinates {(52,0.8)(2450,0.8)};
  \addplot[cfid]  table[x=n,y=fid, col sep=comma]{data/checker256_power.csv};
  \addplot[ccsw]  table[x=n,y=csw, col sep=comma]{data/checker256_power.csv};
  \addplot[ccmmd] table[x=n,y=cmmd,col sep=comma]{data/checker256_power.csv};
  \addplot[cours] table[x=n,y=msw, col sep=comma]{data/checker256_power.csv};
  \node[font=\scriptsize, anchor=south west, inner sep=0, yshift=4pt] at (axis cs:64,1) {MSW (ours)};
  \node[font=\scriptsize, anchor=north west, inner sep=0, xshift=4pt, cbBlue] at (axis cs:1000,0.56) {CMMD};
  \node[font=\scriptsize, anchor=south east, inner sep=0, cbSky!85!black] at (axis cs:900,0.66) {CSW};
  \node[font=\scriptsize, anchor=south east, inner sep=0, yshift=4pt, cbVerm] at (axis cs:2000,0.05) {FID};
  \end{axis}
  \node[rotate=90, inner sep=0, font=\scriptsize] at (ylab |- {0,0}) {one sample};
\end{tikzpicture}
\caption{A red--green checkerboard on ImageNet, $\varepsilon=0.02$.}
\label{fig:checker256}
\end{figure}
 \begin{figure*}[t]
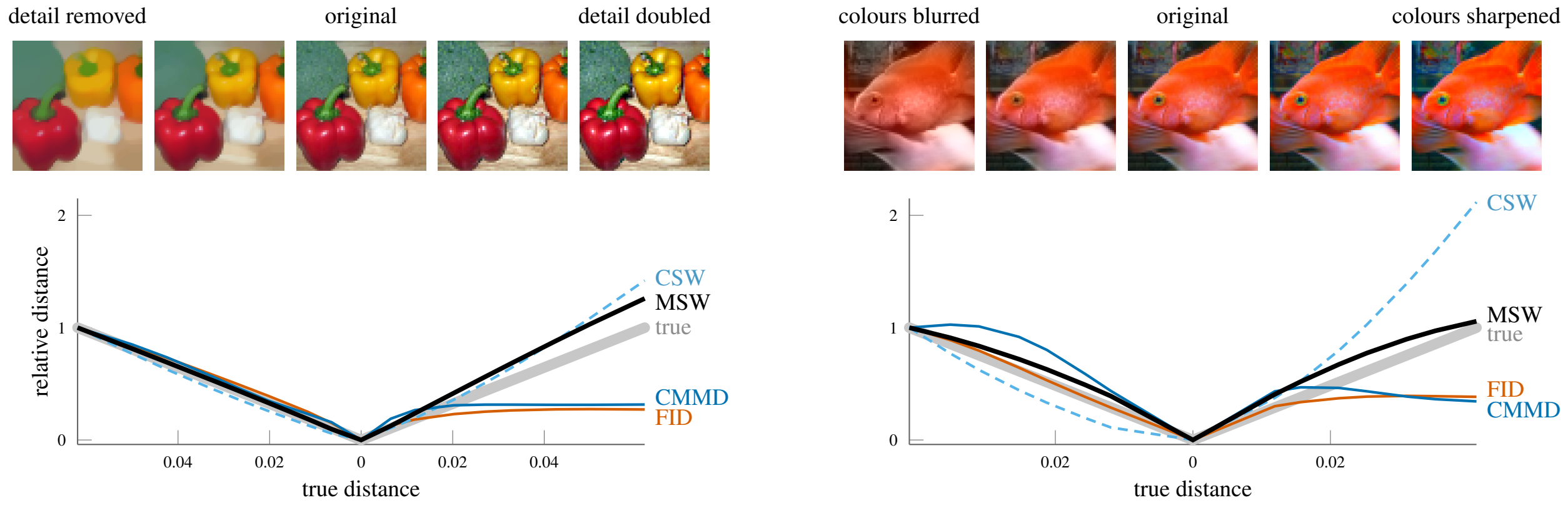

\centering
\begin{tikzpicture}[x=1mm, y=1mm]
  \tikzset{lab/.style={font=\scriptsize, inner sep=0}}
  \foreach \k/\f/\t in {0/m2/detail removed, 1/m1/, 2/0/original, 3/p1/, 4/p2/detail doubled} {
    \node[inner sep=0] at (0+15*\k, 0) {\includegraphics[width=13.7mm, height=13.7mm]{figures/detail_tv_\f.png}};
    \node[inner sep=0, font=\scriptsize, anchor=base] at (0+15*\k, 8.75) {\strut\t};
  }
  \begin{axis}[mswbase, grid=none, anchor=north west, at={(0mm, -9.8mm)}, scale only axis, width=60mm, height=26mm,
    axis x line*=bottom, axis y line*=left, minor tick style={draw=none}, clip=false,
    every axis plot/.append style={no marks},
    xmin=-0.06195, xmax=0.06195, ymin=-0.04, ymax=2.15,
    xtick={-0.04,-0.02,0,0.02,0.04}, xticklabels={0.04,0.02,0,0.02,0.04}, scaled x ticks=false,
    ytick={0,1,2}, yticklabels={0,1,2},
    xlabel={true distance}, ylabel={relative distance}, ylabel style={yshift=-4pt},]
  \addplot[black!22, line width=3.2pt, line cap=round] table[x=x,y=true, col sep=comma]{data/detail_tv.csv};
  \addplot[cfid]  table[x=x,y=fid,  col sep=comma]{data/detail_tv.csv};
  \addplot[ccsw]  table[x=x,y=csw,  col sep=comma]{data/detail_tv.csv};
  \addplot[ccmmd] table[x=x,y=cmmd, col sep=comma]{data/detail_tv.csv};
  \addplot[cours] table[x=x,y=msw,  col sep=comma]{data/detail_tv.csv};
  \node[lab, anchor=west, xshift=3pt, yshift=-2.3pt, cbVerm] at (axis cs:0.06195,0.272) {FID};
  \node[lab, anchor=west, xshift=3pt, yshift=2.3pt, cbBlue] at (axis cs:0.06195,0.316) {CMMD};
  \node[lab, anchor=west, xshift=3pt, yshift=0.0pt, black!45] at (axis cs:0.06195,1.000) {true};
  \node[lab, anchor=west, xshift=3pt, yshift=-1.0pt, black] at (axis cs:0.06195,1.260) {MSW};
  \node[lab, anchor=west, xshift=3pt, yshift=1.0pt, cbSky!85!black] at (axis cs:0.06195,1.419) {CSW};
  \end{axis}

  \foreach \k/\f/\t in {0/m2/colours blurred, 1/m1/, 2/0/original, 3/p1/, 4/p2/colours sharpened} {
    \node[inner sep=0] at (88+15*\k, 0) {\includegraphics[width=13.7mm, height=13.7mm]{figures/detail_colour_\f.png}};
    \node[inner sep=0, font=\scriptsize, anchor=base] at (88+15*\k, 8.75) {\strut\t};
  }
  \begin{axis}[mswbase, grid=none, anchor=north west, at={(88mm, -9.8mm)}, scale only axis, width=60mm, height=26mm,
    axis x line*=bottom, axis y line*=left, minor tick style={draw=none}, clip=false,
    every axis plot/.append style={no marks},
    xmin=-0.04123, xmax=0.04123, ymin=-0.04, ymax=2.15,
    xtick={-0.02,0,0.02}, xticklabels={0.02,0,0.02}, scaled x ticks=false,
    ytick={0,1,2}, yticklabels={0,1,2},
    xlabel={true distance}, ]
  \addplot[black!22, line width=3.2pt, line cap=round] table[x=x,y=true, col sep=comma]{data/detail_colour.csv};
  \addplot[cfid]  table[x=x,y=fid,  col sep=comma]{data/detail_colour.csv};
  \addplot[ccsw]  table[x=x,y=csw,  col sep=comma]{data/detail_colour.csv};
  \addplot[ccmmd] table[x=x,y=cmmd, col sep=comma]{data/detail_colour.csv};
  \addplot[cours] table[x=x,y=msw,  col sep=comma]{data/detail_colour.csv};
  \node[lab, anchor=west, xshift=3pt, yshift=-2.4pt, cbBlue] at (axis cs:0.04123,0.344) {CMMD};
  \node[lab, anchor=west, xshift=3pt, yshift=2.4pt, cbVerm] at (axis cs:0.04123,0.384) {FID};
  \node[lab, anchor=west, xshift=3pt, yshift=-2.1pt, black!45] at (axis cs:0.04123,1.000) {true};
  \node[lab, anchor=west, xshift=3pt, yshift=2.1pt, black] at (axis cs:0.04123,1.057) {MSW};
  \node[lab, anchor=west, xshift=3pt, yshift=0.0pt, cbSky!85!black] at (axis cs:0.04123,2.117) {CSW};
  \end{axis}

\end{tikzpicture}
\caption{Removing or adding detail (left) and colour detail (right) in photographs.}
\label{fig:detail}
\end{figure*}
 We first inject a perturbation of amplitude $\varepsilon$ into one of the two
sets. Table~\ref{tab:more} gives the detection threshold and the sample
requirement of the four metrics for ten perturbations.

\textbf{Gaussian noise.} Gaussian noise is the first disturbance on which FID was
validated~\cite{heusel2017gans}. Score-based generators can also leave in
their images a noise that the eye does not
see~\cite{jolicoeur2021adversarial}. We add Gaussian noise of standard
deviation $\varepsilon$, for pixel values in $[0,1]$, to crops of CelebA
face images~\cite{ref:celeba} (Table~\ref{tab:more}, first row). MSW detects the weakest noise, about an eighth of a
grey level, and it needs the fewest images at a fixed amplitude. This sensitivity comes mainly from the chromatic
slices. The responses of unperturbed faces along the opponent axes are
tightly concentrated, and even faint noise broadens them measurably
(Figure~\ref{fig:marginalsrow}).

\textbf{Checkerboard.} Checkerboard patterns are a common artefact of generated images. They are
left by the upsampling layers of the generator, and they are often visible
only at close range~\cite{odena2016deconvolution}. Such periodic traces
are found across architectures and datasets, and a classifier can use them
to recognize generated images~\cite{frank2020leveraging}. We add $+\varepsilon$ or
$-\varepsilon$ to every pixel in such a pattern, on COCO photographs of
$512$ pixels~\cite{ref:coco}. The pattern is either grey, with the same sign
in the three colour channels, or red--green, with opposite signs in the red
and green channels. MSW detects both patterns at a far smaller amplitude
than the other metrics, and it needs the fewest images
(Table~\ref{tab:more}). The pattern shifts the output of
the highest-frequency filter up at one position and down at the next, so that its
histogram splits in two
(Figure~\ref{fig:marginalsrow}, right).

The networks of FID and CMMD resize their inputs. On ImageNet images of
$256$ pixels~\cite{ref:imagenet}, this removes more than half of the
amplitude of the red--green pattern. In
Figure~\ref{fig:checker256}, MSW detects this pattern in every run from
$128$ images, and CMMD and CSW only with the largest sets. On the same images, CMMD detects a
blur from smaller sets, so the failure is specific to
the pattern (Appendix~\ref{app:protocol}). FID does not detect the pattern
at any set size, because its latent space is nearly invariant to it
(Section~\ref{sec:intro}).

\textbf{Blur, shading and textures.} The other rows of Table~\ref{tab:more} are seven further
perturbations (Appendix~\ref{app:more}). On CelebA and
CIFAR-10~\cite{ref:cifar} images, we apply a
Gaussian blur of standard deviation $\varepsilon$ pixels, the second
disturbance of~\cite{heusel2017gans}. On the COCO photographs, we add a
smooth random shading of standard deviation $\varepsilon$. The
\emph{synthetic textures} are Gaussian
random fields with a power-law spectrum, whose distribution is known
exactly. We change their spectral slope by $\varepsilon$, add white noise of
standard deviation $\varepsilon$, or give them heavier tails with the same
spectrum (a Student-$t$ distribution with $1/\varepsilon$ degrees of
freedom). The last row changes the slope of textures that already have heavy
tails.

MSW has the smallest detection threshold and the smallest sample requirement
in every row except shading. Shading alters only the lowest frequencies,
and CMMD detects it at half the amplitude. The other perturbations are of
different kinds. Blur removes the highest frequencies, and heavier tails
do not change the power spectrum of the textures. Section~\ref{sec:ablation} examines the
ingredients of MSW one by one.

\subsection{Comparison with a true distance}
\label{sec:detail}
Here, the true distance between the two sets is known. Let $T$ be a map
that changes an image, and $T_\sharp\mu$ the distribution of the changed
images. By Brenier's theorem~\cite{brenier1991polar}, if $T$ is the
gradient of a convex function, then it is an optimal transport
map~\cite[Thm.~1.48]{santambrogioOptimalTransportApplied2015}. The coupling
of each image $\V x$ with $T(\V x)$ then attains the minimum
in~\eqref{eq:wp} for $p=2$, and the distance is the root mean square of
the change,
\begin{equation}
\label{eq:exact}
W_2^2(\mu,T_\sharp\mu)=\int\norm{\V x-T(\V x)}^2\,\dd\mu(\V x),
\end{equation}
which we compute from the images. This holds for every denoiser of the form
$T(\V x)=\arg\min_{\V y}\tfrac12\norm{\V x-\V y}^2+f(\V y)$ with $f$
convex~\cite{ref:moreau}. The denoiser removes the detail $\V x-T(\V x)$
from the image. The map $\V x\mapsto\V x+(\V x-T(\V x))$ adds the same
detail instead, and it is also the gradient of a convex function
(Appendix~\ref{app:protocol}). Both maps move each image by the same
amount, and thus give the same distance~\eqref{eq:exact}.

Figure~\ref{fig:detail} shows each metric against this distance, for the
total-variation denoiser~\cite{ref:rof} (left) and for a blur of the colours (right). Each
metric is divided by its value at the left end. The grey line is the true
distance, and a metric proportional to it stays on this line. MSW follows
it on both sides. FID and CMMD grow with it when detail is removed, but
not when it is added. CSW also grows on both sides, but in the right panel it reaches twice the
true distance.

\subsection{Generated images}
\label{sec:sd}
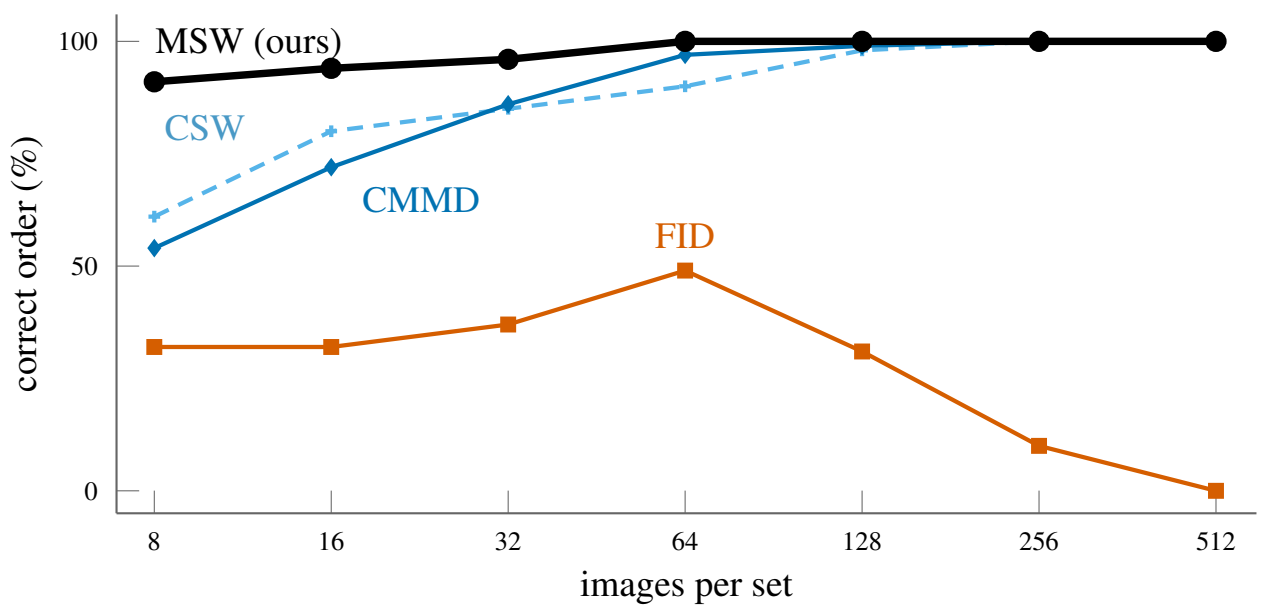
\begin{figure}[t]
\centering
\begin{tikzpicture}[x=1mm, y=1mm]
  \begin{axis}[mswbase, grid=none, scale only axis, width=75.4mm, height=33mm,
    axis x line*=bottom, axis y line*=left,
    xmode=log, xmin=6.9, xmax=600, ymin=-5, ymax=106,
    xtick={8,16,32,64,128,256,512}, xticklabels={8,16,32,64,128,256,512}, minor tick style={draw=none},
    ytick={0,50,100}, yticklabels={0,50,100},
    xlabel={images per set}, ylabel={correct order (\%)}, ylabel style={yshift=-4pt},
    clip=false]
  \path (axis cs:8,100) ++(0mm,3.7mm);
  \addplot[ccsw]  table[x=n,y=csw, col sep=comma]{data/sd_share.csv};
  \addplot[cfid]  table[x=n,y=fid, col sep=comma]{data/sd_share.csv};
  \addplot[ccmmd] table[x=n,y=cmmd,col sep=comma]{data/sd_share.csv};
  \addplot[cours] table[x=n,y=msw, col sep=comma]{data/sd_share.csv};
  \node[font=\scriptsize, inner sep=0, anchor=south west, yshift=4pt] at (axis cs:8,91) {MSW (ours)};
  \node[font=\scriptsize, inner sep=0, anchor=north west, cbBlue] at (axis cs:18,68) {CMMD};
  \node[font=\scriptsize, inner sep=0, anchor=south, yshift=4pt, cbVerm] at (axis cs:64,49) {FID};
  \node[font=\scriptsize, inner sep=0, anchor=north west, cbSky!85!black] at (axis cs:8.3,84) {CSW};
  \end{axis}
\end{tikzpicture}
\caption{Percentage of random draws of $N$ images per set in which a metric
orders the six Stable Diffusion sets correctly.}
\label{fig:sd}
\end{figure}
 The next experiment follows the benchmark of the CMMD
paper~\cite{jayasumana2024rethinking}. Stable Diffusion 1.4 generates one
image for each caption of COCO-30k, and the images after the last six of its
sampling iterations form six sets of increasing quality, which are compared
with the COCO images of the same captions (Appendix~\ref{app:protocol}). On
the full sets, CMMD and MSW decrease at every iteration, whereas FID first
decreases and then increases, the failure reported
in~\cite{jayasumana2024rethinking}. The question here is how many images
each metric needs to recover the order. We draw $N$ captions at random, keep
only their images in the six sets and in COCO, and score the six sets. We
repeat this draw and count how often a metric orders all six sets correctly
(Figure~\ref{fig:sd}). MSW recovers the order most reliably, in about nine
draws out of ten from the smallest size and in every draw from $64$ images.
CMMD and CSW catch up only at the larger sizes. FID recovers the order in fewer
than half of the draws at every size and in none at the largest, where it
reproduces its wrong order. The
value of MSW is also stable in $N$. On a thirtieth of the images, FID reads
several times its value on the full sets, whereas CMMD and MSW stay within a
few percent of theirs (Appendix~\ref{app:protocol}).

\subsection{Super-resolution}
\label{sec:restoration}
\begin{figure}[t]
\centering
\begin{tikzpicture}[x=1mm, y=1mm]
  \pgfplotsset{srpanel/.style={mswbase, grid=none, anchor=north west, scale only axis, width=31.2mm, height=27.3mm,
    axis x line*=bottom, axis y line*=left, minor tick style={draw=none},
    every axis plot/.append style={no marks}, title style={font=\scriptsize, yshift=-3pt}}}
  \tikzset{lab/.style={font=\scriptsize, inner sep=0}}
  \foreach \k/\f/\t/\s in {0/real/original/, 1/a000/$\alpha=0$/PSNR-trained, 2/a100/$\alpha=1$/GAN-trained,
                           3/a140/$\alpha=1.4$/} {
    \node[inner sep=0] at (19.37*\k, 0) {\includegraphics[width=17.3mm, height=17.3mm]{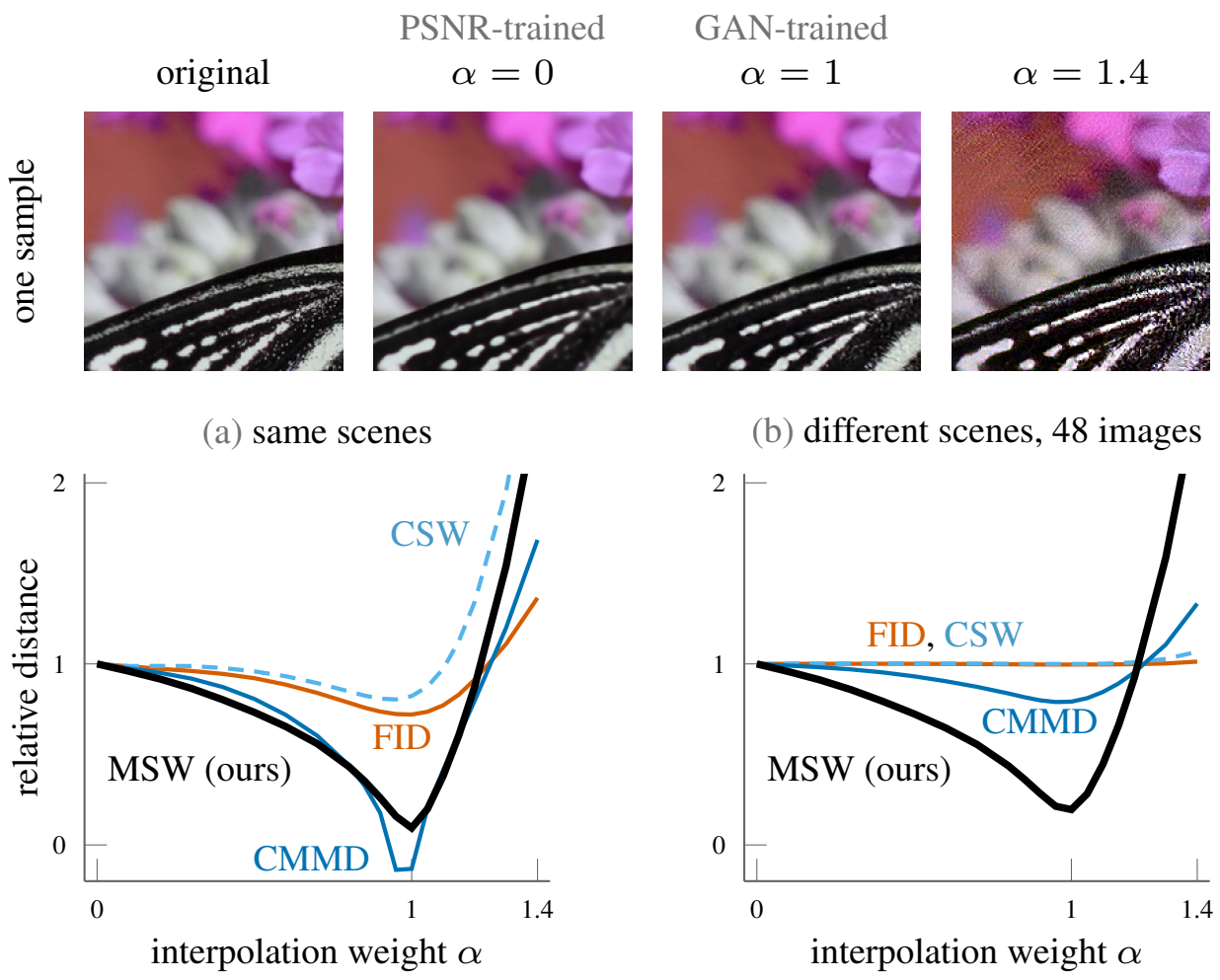}};
    \node[inner sep=0, font=\scriptsize, anchor=base] at (19.37*\k, 10.55) {\strut\t};
    \node[inner sep=0, font=\scriptsize, anchor=base, black!55] at (19.37*\k, 13.5) {\strut\s};
  }
  \begin{axis}[srpanel, name=a, at={(-8.65mm, -15.6mm)},
    xmin=-0.04, xmax=1.44, ymin=-0.2, ymax=2.05, clip=true,
    xtick={0,1,1.4}, xticklabels={0,1,1.4}, ytick={0,1,2}, yticklabels={0,1,2},
    title={{\color{black!55}(a)} same scenes},
    xlabel={interpolation weight $\alpha$}, ylabel={relative distance}, ylabel style={yshift=-4pt, name=ylab}]
  \addplot[cfid]  table[x=alpha,y=fid,  col sep=comma]{data/restoration_path.csv};
  \addplot[ccsw]  table[x=alpha,y=csw,  col sep=comma]{data/restoration_path.csv};
  \addplot[ccmmd] table[x=alpha,y=cmmd, col sep=comma]{data/restoration_path.csv};
  \addplot[cours] table[x=alpha,y=msw,  col sep=comma]{data/restoration_path.csv};
  \node[lab, anchor=north west] at (axis cs:0.03,0.5) {MSW (ours)};
  \node[lab, anchor=south east, cbBlue] at (axis cs:0.87,-0.15) {CMMD};
  \node[lab, anchor=north, cbVerm] at (axis cs:0.97,0.67) {FID};
  \node[lab, anchor=east, cbSky!85!black] at (axis cs:1.19,1.72) {CSW};
  \end{axis}
  \begin{axis}[srpanel, name=b, at={(35.56mm, -15.6mm)},
    xmin=-0.04, xmax=1.44, ymin=-0.2, ymax=2.05, clip=true,
    xtick={0,1,1.4}, xticklabels={0,1,1.4}, ytick={0,1,2}, yticklabels={0,1,2},
    title={{\color{black!55}(b)} different scenes, 48 images},
    xlabel={interpolation weight $\alpha$}]
  \addplot[cfid]  table[x=alpha,y=fid,  col sep=comma]{data/restoration_path_d12.csv};
  \addplot[ccsw]  table[x=alpha,y=csw,  col sep=comma]{data/restoration_path_d12.csv};
  \addplot[ccmmd] table[x=alpha,y=cmmd, col sep=comma]{data/restoration_path_d12.csv};
  \addplot[cours] table[x=alpha,y=msw,  col sep=comma]{data/restoration_path_d12.csv};
  \node[lab, anchor=north west] at (axis cs:0.03,0.5) {MSW (ours)};
  \node[lab, anchor=north, cbBlue] at (axis cs:0.9,0.76) {CMMD};
  \node[lab, anchor=south] at (axis cs:0.6,1.06) {{\color{cbVerm}FID}, {\color{cbSky!85!black}CSW}};
  \end{axis}
  \node[rotate=90, inner sep=0, font=\scriptsize] at (ylab |- {0,0}) {one sample};
\end{tikzpicture}
\caption{Super-resolution along the ESRGAN interpolation path.}
\label{fig:restoration}
\end{figure}
 In this experiment, the order that a metric should recover is known from
the literature rather than chosen by us. A super-resolution network trained
to minimize the pixel error (\emph{PSNR-trained}) produces faithful but
blurry images. When the same network is trained further with an adversarial
loss (\emph{GAN-trained}), its outputs are less faithful pixel by pixel, but
they are sharp and their distribution is closer to that of real images. This
is the perception--distortion trade-off~\cite{ref:pdtradeoff}.
ESRGAN~\cite{ref:esrgan} interpolates linearly between the weights of the two
networks, with a weight $\alpha$ on the GAN-trained one. The outputs go from
blurry at $\alpha=0$ to sharp at $\alpha=1$, and they are over-sharpened
beyond (Figure~\ref{fig:restoration}, top). The distance between the outputs
and the originals should therefore decrease up to about $\alpha=1$ and
increase beyond.

We enlarge the DIV2K validation images~\cite{ref:div2k} by a factor of four
(Appendix~\ref{app:protocol}).
In Figure~\ref{fig:restoration}(a), the outputs are compared with the
originals of the same scenes, and each metric is divided by its value at
$\alpha=0$. All four metrics are smallest near the GAN-trained network. MSW
follows CMMD closely, without a pretrained network, whereas FID and CSW have
a shallow minimum.

In practice, two sets rarely show the same scenes. In
Figure~\ref{fig:restoration}(b), the outputs and the originals come from
disjoint sets of $48$ images, and each curve is the mean over $200$ random
draws. MSW keeps a clear minimum at the GAN-trained network. FID and CSW are
now flat. CMMD decreases less than MSW, although it ranks the
two networks correctly in more draws (Appendix~\ref{app:protocol}). Four
other pairs of networks lead to the same conclusions, with the minimum of
MSW slightly beyond $\alpha=1$ for two of them.

\subsection{Semantic changes}
\label{sec:semantic}
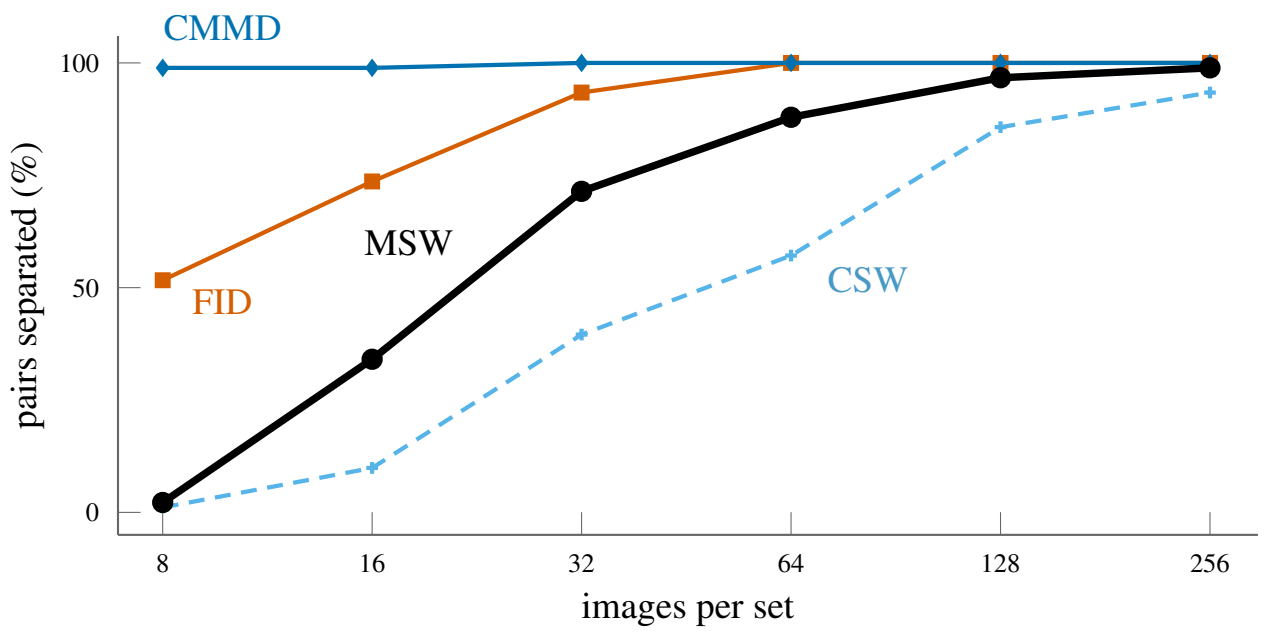
\begin{figure}[t]
\centering
\begin{tikzpicture}[x=1mm, y=1mm]
  \begin{axis}[mswbase, grid=none, scale only axis, width=75.4mm, height=33mm,
    axis x line*=bottom, axis y line*=left,
    xmode=log, xmin=6.9, xmax=300, ymin=-5, ymax=106,
    xtick={8,16,32,64,128,256}, xticklabels={8,16,32,64,128,256}, minor tick style={draw=none},
    ytick={0,50,100}, yticklabels={0,50,100},
    xlabel={images per set}, ylabel={pairs separated (\%)}, ylabel style={yshift=-4pt},
    clip=false]
  \path (axis cs:8,100) ++(0mm,3.7mm);
  \addplot[ccsw]  table[x=n,y=csw, col sep=comma]{data/classes_share_coco.csv};
  \addplot[cfid]  table[x=n,y=fid, col sep=comma]{data/classes_share_coco.csv};
  \addplot[ccmmd] table[x=n,y=cmmd,col sep=comma]{data/classes_share_coco.csv};
  \addplot[cours] table[x=n,y=msw, col sep=comma]{data/classes_share_coco.csv};
  \node[font=\scriptsize, inner sep=0, anchor=south west, yshift=4pt, cbBlue] at (axis cs:8,100) {CMMD};
  \node[font=\scriptsize, inner sep=0, anchor=north west, cbVerm] at (axis cs:8.8,50) {FID};
  \node[font=\scriptsize, inner sep=0, anchor=south east] at (axis cs:21,57) {MSW};
  \node[font=\scriptsize, inner sep=0, anchor=north west, cbSky!85!black] at (axis cs:72,55) {CSW};
  \end{axis}
\end{tikzpicture}
\caption{Percentage of the pairs of COCO classes that a metric separates from $N$
images per set.}
\label{fig:classes}
\end{figure}

MSW contains no learned component, whereas the networks of FID and CMMD
are trained to classify images or to match them with text. A change of
semantic content is thus the hardest case for MSW, and we expect FID and
CMMD to be ahead. We compare the metrics on pairs of COCO classes. A
photograph belongs to the class of its largest annotated object, and the
$14$ classes with the most photographs form $91$ pairs.
For each pair, the two sets contain $N$ images of one
class and $N$ images of the other, and the null distribution
of each baseline is measured on two disjoint sets of the first class. The
detection rule is that of Section~\ref{sec:setup}.
Figure~\ref{fig:classes} shows the share of pairs
separated with power $0.8$, and Table~\ref{tab:classpairs} lists them
(Appendix~\ref{app:protocol}).
 
MSW separates most pairs from a few dozen images per set, and all but one
from the largest sets. As expected, FID and CMMD need several times fewer
images. CSW needs more than MSW. The pair that MSW does not separate is car against truck, and
the other hard pairs are mostly two vehicles, such as
bus against truck. Image
categories differ in their low-level statistics, since the power spectra of
natural images depend on the scene category~\cite{ref:torralba2003} and a
classifier restricted to small patches reaches a high accuracy on
ImageNet~\cite{ref:bagnet}. Accumulated over all positions of the images,
these differences separate nearly every pair of classes.

\subsection{Ablation}
\label{sec:ablation}
\begin{figure}[t]
\centering
\pgfplotsset{ablax/.style={mswbase, grid=none, anchor=north west, scale only axis, width=20.6mm, height=16.8mm,
  axis x line*=bottom, axis y line*=left, minor tick style={draw=none}, clip=false,
  ymin=-0.05, ymax=1.06, ytick={0,0.8,1}, yticklabels={},
  title style={font=\scriptsize, yshift=-3pt, black!55, text height=1.5ex, text depth=0.5ex},
  every axis plot/.append style={mark size=0.9pt}},
  cvar/.style={black!45, line width=0.9pt, mark=*, mark size=0.9pt}}
\begin{tikzpicture}[x=1mm, y=1mm]
  \begin{axis}[ablax, xmode=log, at={(0mm, 0mm)}, xmin=0.00009, xmax=0.007,
    xtick={0.0005,0.005}, xticklabels={0.0005,0.005}, yticklabels={0,0.8,1},
    title={random filters}, xlabel={noise $\varepsilon$}, ylabel={power}, ylabel style={yshift=-4pt}]
  \addplot[thr] coordinates {(0.00009,0.8)(0.007,0.8)};
  \addplot[cvar]  table[x=eps,y=random, col sep=comma]{data/ablation_filters.csv};
  \addplot[cours, mark size=1.1pt] table[x=eps,y=msw, col sep=comma]{data/ablation_filters.csv};
  \end{axis}
  \begin{axis}[ablax, xmode=log, at={(27.2mm, 0mm)}, xmin=0.00009, xmax=0.007,
    xtick={0.0005,0.005}, xticklabels={0.0005,0.005},
    title={R, G, B axes}, xlabel={noise $\varepsilon$}]
  \addplot[thr] coordinates {(0.00009,0.8)(0.007,0.8)};
  \addplot[cvar]  table[x=eps,y=rgb, col sep=comma, restrict x to domain=-9.3:-5.2]{data/ablation_colour.csv};
  \addplot[cours, mark size=1.1pt] table[x=eps,y=msw, col sep=comma, restrict x to domain=-9.3:-5.2]{data/ablation_colour.csv};
  \end{axis}
  \begin{axis}[ablax, xmode=log, at={(54.4mm, 0mm)}, xmin=0.015, xmax=0.9,
    xtick={0.08,0.32}, xticklabels={0.08,0.32},
    title={one scale}, xlabel={low frequencies $\varepsilon$}]
  \addplot[thr] coordinates {(0.015,0.8)(0.9,0.8)};
  \addplot[cvar]  table[x=eps,y=one, col sep=comma]{data/ablation_scales.csv};
  \addplot[cours, mark size=1.1pt] table[x=eps,y=msw, col sep=comma]{data/ablation_scales.csv};
  \end{axis}
\end{tikzpicture}
\caption{MSW (black) with one ingredient replaced (grey).}
\label{fig:ablation}
\end{figure}
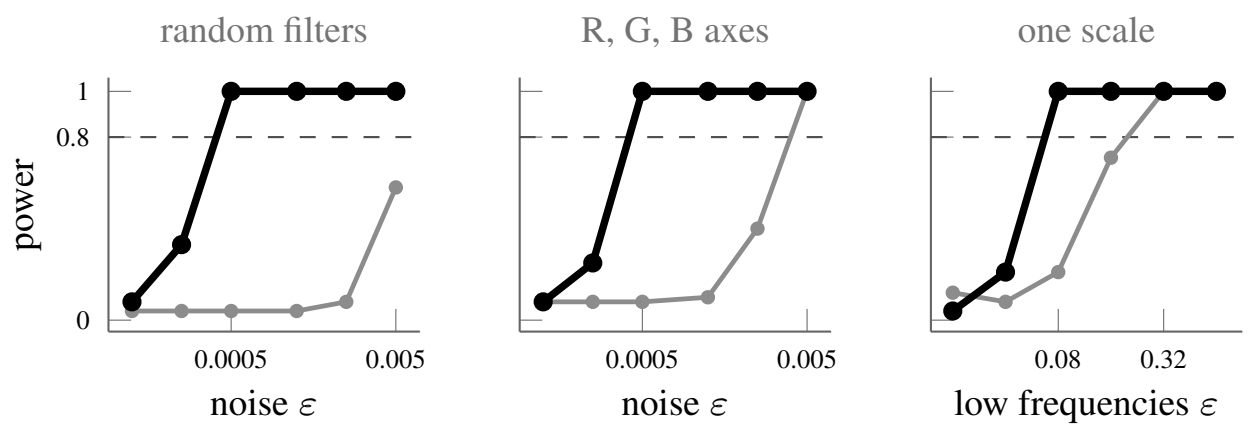
 In Figure~\ref{fig:ablation}, we replace the DCT atoms, the opponent
colour axes and the scales of MSW one at a time. With random
filters, Gaussian noise on CelebA images is not detected at ten times the
threshold of MSW. With learned filters, the thresholds do not change
(Appendix~\ref{app:estimation}). With the R, G, B axes, the threshold of
this noise is ten times larger. With one scale, the threshold of a change
in the lowest frequencies of a synthetic texture is four times larger
(Appendix~\ref{app:more}).

\section{Limitations}
\label{sec:scope}
MSW has the following limitations.

\textbf{Global colour offsets.} All filters have zero mean, so
that MSW is invariant to a constant offset of the channels, at any
amplitude. CMMD detects such an offset (Appendix~\ref{app:estimation}).

\textbf{Dependence between filters.} MSW compares the responses of each
filter separately. It does not compare how the responses of different
filters vary together, although they are not
independent~\cite{simoncelli2001natural}.

\textbf{Semantic changes.} Latent-based metrics separate pairs of COCO
classes with several times fewer images than MSW (Section~\ref{sec:semantic}).

\textbf{Memorization and definiteness.} Like FID, MSW ignores the identity of
individual images and therefore does not detect a generator that reproduces
its training set. The bank also spans finitely many frequency bands, and
two distributions that agree on all of them are indistinguishable.

\section{Conclusion}
In most of our experiments, MSW outperforms FID, CMMD and CSW. It
detects smaller perturbations from fewer images, and it follows a true
distance more closely. Latent-based metrics still need fewer images for
the changes that their networks are trained to recognize, such as a change
of object class. MSW has no trained component. Its design rests on
well-studied properties of natural images: their stationarity, their
decorrelation by fixed filters and their similar statistics at every
scale. Our results show that these properties are enough to build a
simple and interpretable metric that competes with large pretrained
networks.

\newpage
\section*{LLM Usage Statement}
We used an LLM assistant (Claude, Anthropic), mainly to refine
author-produced content: the code and parts of the text. The authors
verified all content and take full responsibility for it.

{
    \small
    \bibliographystyle{ieeenat_fullname_unsrt}

}

\clearpage
\maketitlesupplementary

\renewcommand{\thefigure}{S\arabic{figure}}
\renewcommand{\thetable}{S\arabic{table}}
\renewcommand{\theequation}{S\arabic{equation}}
\setcounter{figure}{0}\setcounter{table}{0}\setcounter{equation}{0}
\renewcommand{\thelemma}{S\arabic{lemma}}
\renewcommand{\theproposition}{S\arabic{proposition}}

\appendix
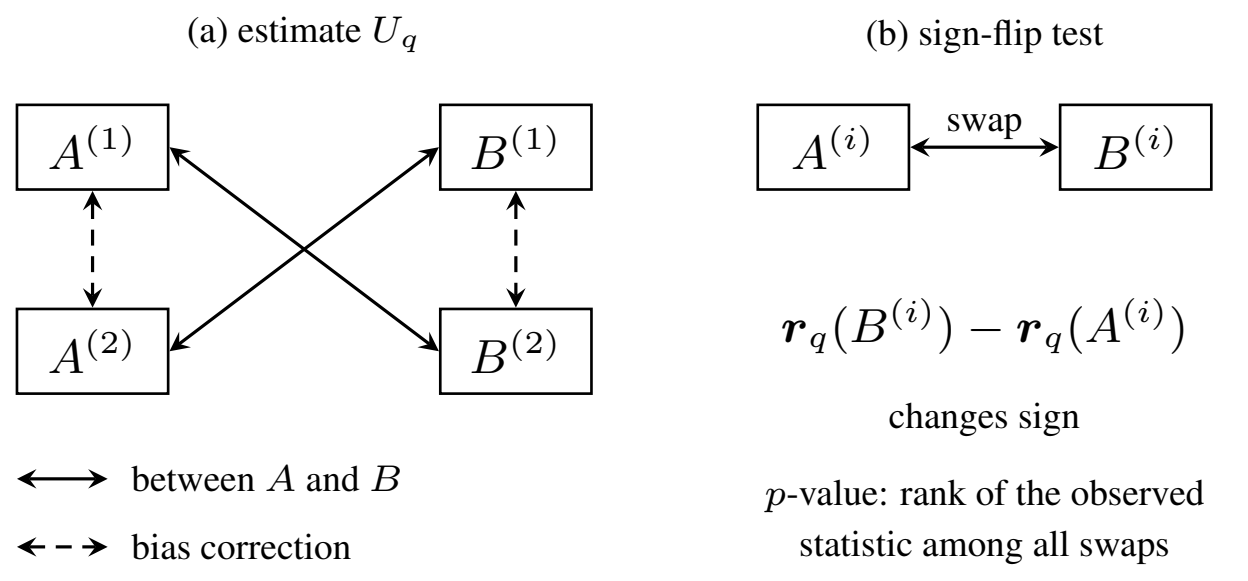
\begin{figure}[t]
\centering
\begin{tikzpicture}[x=1mm, y=1.125mm, >=stealth,
  blk/.style={draw, line width=0.5pt, minimum width=10mm, minimum height=5.6mm, inner sep=0, font=\small},
  lab/.style={font=\scriptsize, inner sep=0},
  btw/.style={<->, line width=0.6pt},
  wth/.style={<->, line width=0.6pt, densely dashed}]
  \node[lab, anchor=south] at (19, 20.5) {(a) estimate $U_q$};
  \node[blk] (a1) at (5, 15) {$A^{(1)}$};
  \node[blk] (a2) at (5, 3)  {$A^{(2)}$};
  \node[blk] (b1) at (33, 15) {$B^{(1)}$};
  \node[blk] (b2) at (33, 3)  {$B^{(2)}$};
  \draw[btw] (a1.east) -- (b2.west);
  \draw[btw] (a2.east) -- (b1.west);
  \draw[wth] (a1.south) -- (a2.north);
  \draw[wth] (b1.south) -- (b2.north);
  \draw[btw] (0, -4.5) -- (6, -4.5);
  \node[lab, anchor=west] at (7.5, -4.5) {between $A$ and $B$};
  \draw[wth] (0, -8.5) -- (6, -8.5);
  \node[lab, anchor=west] at (7.5, -8.5) {bias correction};
  \node[lab, anchor=south] at (64, 20.5) {(b) sign-flip test};
  \node[blk] (ai) at (54, 15) {$A^{(i)}$};
  \node[blk] (bi) at (74, 15) {$B^{(i)}$};
  \draw[<->, line width=0.6pt] (ai.east) -- node[lab, above=1.2pt] {swap} (bi.west);
  \node[font=\small, inner sep=0] at (64, 4.5) {$\V r_q(B^{(i)})-\V r_q(A^{(i)})$};
  \node[lab] at (64, -1) {changes sign};
  \node[lab] at (64, -5.5) {$p$-value: rank of the observed};
  \node[lab] at (64, -8.5) {statistic among all swaps};
\end{tikzpicture}
\caption{Blocks of the estimate $U_q$ (a) and of the exact test (b).}
\label{fig:blocks}
\end{figure}
 \section{Computational details}
\label{app:estimation}
We will define hereafter the practical way of computing MSW with finite samples and the exact test that we use
for the detection of perturbations.
In the continuum, recall~\eqref{eq:msw} that
\begin{equation*}
\mathrm{MSW}^2(\mu,\nu)=\sum_q\lambda_q\,W_2^2\big(\mu_q,\nu_q\big),
\end{equation*}
where $\mu_q$ and $\nu_q$ are the slice distributions of $\mu$ and $\nu$.
\paragraph{Bias due to finite-sample sets.} For a set of images $A$ and a slice $q$, let
$A_q$ denote the empirical distribution of the $n$ responses
of the slice at all positions of all images of $A$. It replaces $\mu_q$. Let $\V r_q(A)\in\R^n$ be these responses in increasing order. For
two sets $A$ and $B$ of the same size, we have
$W_2^2(A_q,B_q)=\tfrac{1}{n}\norm{\V r_q(A)-\V r_q(B)}^2$. For two sets drawn from the same distribution,
$W_2^2(A_q,B_q)$ is strictly positive with probability one (the sets of images are independent), and its expectation
decreases only slowly with the number of images. FID has the same property, with a bias of order $1/N$ for sets of $N$ images~\cite{chong2020effectively}. We need to correct for this upward bias.

\paragraph{Bias correction.} Under the null hypothesis (when the two sets of images come from the same distribution), we want our estimator of $\mathrm{MSW}^2(\mu,\nu)$ to be zero on average. 
We split each set by image into two blocks, $A^{(1)},A^{(2)}$ and $B^{(1)},B^{(2)}$ (Figure~\ref{fig:blocks}), and we define,
for the slice $q$, the U-statistic
\begin{equation}
\label{eq:floor}
\begin{aligned}
U_q={}&\tfrac12\underbrace{\big[W_2^2(A_q^{(1)},B_q^{(2)})+W_2^2(A_q^{(2)},B_q^{(1)})\big]}_{\text{between $A$ and $B$}}\\
 &-\tfrac12\underbrace{\big[W_2^2(A_q^{(1)},A_q^{(2)})+W_2^2(B_q^{(1)},B_q^{(2)})\big]}_{\text{bias correction}}.
\end{aligned}
\end{equation}
With $\V d_q^{(i)}=\V r_q(B^{(i)})-\V r_q(A^{(i)})$, we obtain
\begin{equation}
\label{eq:ident}
U_q=\frac{\langle\V d_q^{(1)},\V d_q^{(2)}\rangle}{n}.
\end{equation}
\begin{proposition}
\label{prop:floor}
If all images of $A$ and $B$ are drawn independently from the same
distribution $\mu$, then $\E[U_q]=0$ for every slice $q$ and every $N$.
\end{proposition}
\begin{proof}
The four blocks are identically distributed, so that
$\E[\V d_q^{(1)}]=\E[\V d_q^{(2)}]=\boldsymbol 0$. The two factors of~\eqref{eq:ident}
involve disjoint blocks and are therefore independent, which implies that
$\E[U_q]=\langle\E[\V d_q^{(1)}],\E[\V d_q^{(2)}]\rangle/n=0$.
\end{proof}

In order to improve the accuracy of $U_q$, each set is cut into $L=16$ blocks\footnote{$L=8$
when $N<16$.} of
$\lfloor N/L\rfloor$ images, denoted by $A^{(i)}$ and $B^{(i)}$ for $i=1,\dots,L$. We then take
for $U_q$ the average of $\langle\V d_q^{(i)},\V d_q^{(j)}\rangle/n$ over all pairs
$i\neq j$, which also has zero expectation when $\mu=\nu$.
In practice, we estimate $\mathrm{MSW}^2(\mu,\nu)$ with finite samples through the weighted sum
\begin{equation*}
\mathrm{MSW}^2(A,B)=\sum_q\lambda_qU_q.
\end{equation*}

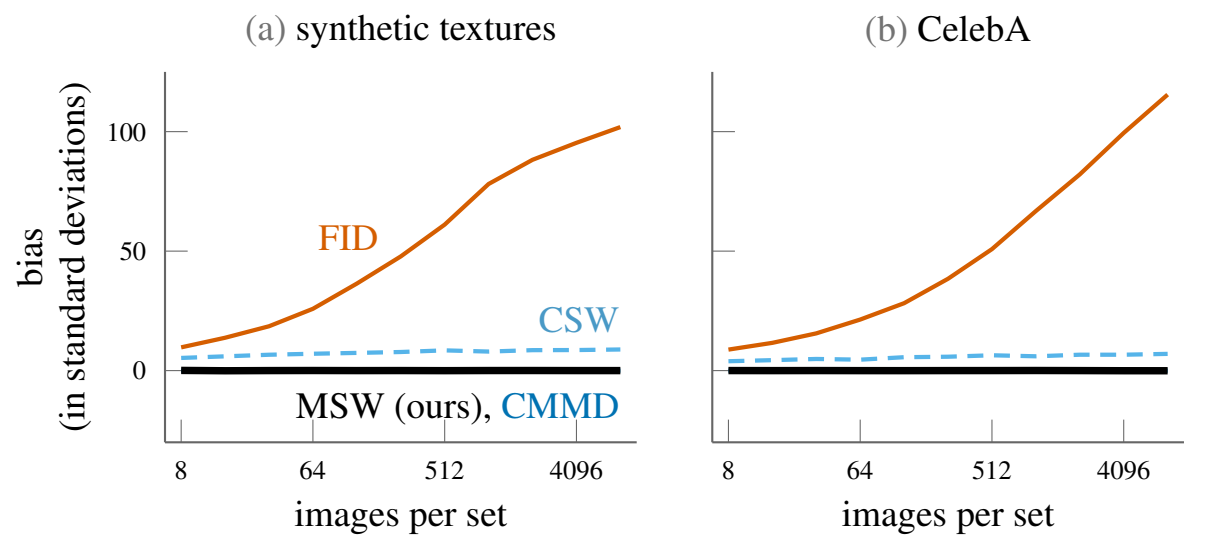
\begin{figure}[t]
\centering
\begin{tikzpicture}[x=1mm, y=1mm]
  \pgfplotsset{nullpanel/.style={mswbase, grid=none, anchor=north west, scale only axis, width=31.2mm, height=24.5mm,
    axis x line*=bottom, axis y line*=left, minor tick style={draw=none},
    every axis plot/.append style={no marks}, title style={font=\scriptsize, yshift=-3pt},
    xmin=6.2, xmax=10600, ymin=-30, ymax=125,
    xtick={8,64,512,4096}, xticklabels={8,64,512,4096},
    ytick={0,50,100}, yticklabels={0,50,100},
    xlabel={images per set}}}
  \tikzset{lab/.style={font=\scriptsize, inner sep=0}}
  \begin{axis}[nullpanel, name=a, at={(0mm, 0mm)}, xmode=log,
    title={{\color{black!55}(a)} synthetic textures},
    ylabel={bias\\(in standard deviations)}, ylabel style={yshift=-4pt, align=center}]
  \addplot[cfid]  table[x=n,y=fid, col sep=comma]{data/null_all_fields.csv};
  \addplot[ccsw]  table[x=n,y=csw, col sep=comma]{data/null_all_fields.csv};
  \addplot[ccmmd] table[x=n,y=cmmd,col sep=comma]{data/null_all_fields.csv};
  \addplot[cours] table[x=n,y=msw, col sep=comma]{data/null_all_fields.csv};
  \node[lab, anchor=south east, cbVerm, xshift=-4pt, yshift=1pt] at (axis cs:256,47.78) {FID};
  \node[lab, anchor=south east, yshift=3pt, cbSky!85!black] at (axis cs:8192,8.84) {CSW};
  \node[lab, anchor=north east, yshift=-4pt] at (axis cs:8192,0) {MSW (ours), {\color{cbBlue}CMMD}};
  \end{axis}
  \begin{axis}[nullpanel, name=b, at={(36.2mm, 0mm)}, xmode=log,
    title={{\color{black!55}(b)} CelebA}, yticklabels={}]
  \addplot[cfid]  table[x=n,y=fid, col sep=comma]{data/null_all_celeba.csv};
  \addplot[ccsw]  table[x=n,y=csw, col sep=comma]{data/null_all_celeba.csv};
  \addplot[ccmmd] table[x=n,y=cmmd,col sep=comma]{data/null_all_celeba.csv};
  \addplot[cours] table[x=n,y=msw, col sep=comma]{data/null_all_celeba.csv};
  \end{axis}
\end{tikzpicture}
\caption{Bias between two sets of the same distribution.}
\label{fig:null}
\end{figure}

Figure~\ref{fig:null} shows the mean of a metric between two sets drawn
from the same distribution, in units of its standard deviation
over $1000$ runs. A run splits $2N$ images taken at random from a pool of
$65{,}536$ synthetic textures or CelebA crops of $32$ pixels, and the four
metrics share the draws. MSW and CMMD stay within $0.1$
standard deviation of zero for all $N$ between $8$ and $8192$. CSW lies $4$
to $9$ standard deviations above zero, and FID $9$ to $115$.
 
\paragraph{Comparison with $W_2$.} Write
$\mu_q=\frac{1}{n_s}\sum_{\V k}(\phi_{q,\V k})_\sharp\mu$, where
$\phi_{q,\V k}(\V x)=\Phi_q(\V x)[\V k]$ is defined in~\eqref{eq:slice}.
The map $\phi_{q,\V k}$ is a linear functional of unit norm, and hence $1$-Lipschitz.\pagebreak{}
In addition,
$W_2^2$ is jointly convex under mixtures with common weights.
We therefore have that
\begin{equation}
\begin{aligned}
W_2^2(\mu_q,\nu_q)&\underset{\makebox[8.3mm]{\scriptsize convexity}}{\le}\frac{1}{n_s}\sum_{\V k}
W_2^2\big((\phi_{q,\V k})_\sharp\mu,(\phi_{q,\V k})_\sharp\nu\big)\\
&\underset{\makebox[8.3mm]{\scriptsize $1$-Lipschitz}}{\le} W_2^2(\mu,\nu).
\end{aligned}
\end{equation}
As the weights $\lambda_q$ are nonnegative and sum to one,
\begin{equation}
    \mathrm{MSW}^2(\mu,\nu)\le\max_q W_2^2(\mu_q,\nu_q)\le W_2^2(\mu,\nu).
\end{equation}

\paragraph{The test for detection.} In order to detect whether $A$ and $B$ come from the same distribution, we perform the classical \textit{sign-flip test}~\cite{lehmann2005testing}.
A choice of swaps is a vector $\V\xi=(\xi_1,\dots,\xi_L)$, where $\xi_i=-1$ if the blocks $A^{(i)}$ and $B^{(i)}$ are exchanged and $\xi_i=1$ otherwise.
For each choice $\V\xi$, we compute a statistic $T^{\V\xi}$ from the blocks after the swaps.
We write $T$ for its value without any swap.
If the two sets come from the same distribution, then the swaps do not change the distribution of the data, and $T$ behaves like any of the $T^{\V\xi}$.
The $p$-value is therefore the fraction of the choices $\V\xi$ for which $T^{\V\xi}$ is larger than $T$.
We use $4096$ choices when $L=16$.\footnote{$256$ choices when $L=8$.}
A swap only changes the sign of $\V d_q^{(i)}$ (Figure~\ref{fig:blocks}), which multiplies
$\langle\V d_q^{(i)},\V d_q^{(j)}\rangle$ by $\xi_i\xi_j$. The swaps therefore need no new convolutions or sorts.

The statistic is the largest of six values $T_1^{\V\xi},\dots,T_6^{\V\xi}$,
\begin{equation}
T^{\V\xi}=\max_{m}\ T_m^{\V\xi}/\rho_m,
\end{equation}
where $\rho_m^2$ is the mean of $(T_m^{\V\xi})^2$ over all the choices $\V\xi$. It brings the six values to the same scale.
Taking the maximum of several statistics over the same swaps is
classical~\cite{westfall1993resampling} and used by recent kernel
tests~\cite{schrab2023mmdagg,biggs2023mmdfuse}.

Each $T_m^{\V\xi}$ is built like $\mathrm{MSW}^2(A,B)$, from the estimates $U_q^{\V\xi}$ computed after the swaps $\V\xi$. It is the weighting that changes.
The fixed weights $\lambda_q$ of~\eqref{eq:msw} make $\mathrm{MSW}^2(A,B)$ comparable from one experiment to the next.
For detection, a slice with a precise estimate should count more.
This is the classical inverse-variance weighting~\cite{cochran1954combination}. We use it with two variances,
\begin{equation}
T_{\mathrm{blocks}}^{\V\xi}=\frac{\sum_q U_q^{\V\xi}/v_q^{\V\xi}}{\sum_q 1/v_q^{\V\xi}},
\quad
T_{\mathrm{swaps}}^{\V\xi}=\frac{\sum_q U_q^{\V\xi}/v'_q}{\sum_q 1/v'_q}.
\end{equation}
The variance $v_q^{\V\xi}$ comes from the $L$ blocks after the swaps $\V\xi$.\footnote{Let $R_{q,i}^{\V\xi}$ denote the average of $\xi_i\xi_j\langle\V d_q^{(i)},\V d_q^{(j)}\rangle/n$ over
$j\neq i$. Then $U_q^{\V\xi}$ is the mean of $R_{q,1}^{\V\xi},\dots,R_{q,L}^{\V\xi}$, and $v_q^{\V\xi}$ is their empirical variance divided by $L$.}
It changes with $\V\xi$.
The variance $v'_q$ is the mean of $(U_q^{\V\xi})^2$ over all the choices $\V\xi$~\cite{biggs2023mmdfuse}.
It is the same for every $\V\xi$.
A third value keeps a single scale,
\begin{equation}
T_{\mathrm{scale}}^{\V\xi}=\max_{s}\ \overline{U}_s^{\V\xi}/\sigma_s^{\V\xi},
\end{equation}
where $\overline{U}_s^{\V\xi}$ is the mean of $U_q^{\V\xi}$ over the slices of the scale $s$ and $\sigma_s^{\V\xi}$ is
its standard error over the blocks. A difference at one scale is then not averaged out by the other scales.
We compute these three values with the red, green and blue
channels, and again with the opponent axes of
Section~\ref{sec:bank} (without the luminance axis for the first two), which gives the six values.
Single-channel images give three values. We check the test on two sets drawn from one distribution, where it should reject in $5\,\%$ of the draws.
On $1000$ draws for each of five image families and for $N=16$ and
$N=500$, it rejects in $4.6\,\%$ to $6.0\,\%$ of them, and in $533$ of $10000$ overall.

\paragraph{Other filters.} In Table~\ref{tab:ablation}, we replace the DCT
atoms by random filters, or by filters learned from image patches by
principal component analysis (PCA), and keep everything else. The random
filters have Gaussian entries and unit norm, and the PCA filters are fitted
to patches of CIFAR-10 images. The random filters are behind in every row
except shading. The PCA filters reach the same thresholds as the DCT atoms,
which need no computation.
\begin{table}[t]
\centering\small
\caption{Threshold $\varepsilon_{0.8}$ with other filters (--: no detection).}
\label{tab:ablation}
\resizebox{\columnwidth}{!}{\begin{tabular}{llccc}
\toprule
 & & DCT & random & PCA \\
\midrule
CelebA & Gaussian noise & 0.0005 & -- & 0.0005 \\
 & Gaussian blur & 0.10 & 0.33 & 0.10 \\
 & shading & 0.08 & 0.08 & 0.08 \\
\addlinespace
CIFAR-10 & checkerboard & 0.01 & -- & 0.01 \\
\addlinespace
synthetic textures & spectral slope & 0.008 & 0.016 & 0.008 \\
\bottomrule
\end{tabular}}
\end{table}

\paragraph{Global colour offsets.} A constant chromatic offset
$\varepsilon\,(1,-1,0)/\sqrt2$ on CelebA crops is invisible to every filter
of the bank, since the atoms have zero mean. The test does not
detect it at any amplitude of the grid. CMMD detects the offset from $0.02$, and
FID from $0.08$.

\section{Protocol details}
\label{app:protocol}
This appendix gives the set-up of the experiments of
Section~\ref{sec:experiments}.
\paragraph{Baselines.} FID uses the Inception-v3 network with the weights and the
preprocessing of the \texttt{pytorch-fid} \mbox{package}~\cite{heusel2017gans}.
CMMD uses the CLIP ViT-L/14 network~\cite{ref:clip} on images resized to $224$ pixels, a Gaussian kernel of bandwidth
$10$~\cite{jayasumana2024rethinking} and the unbiased estimate of the MMD.
In Section~\ref{sec:sd}, we follow its released code, with the $336$-pixel network and the biased estimate.
CSW uses the convolution slicers and the estimate of its released code~\cite{nguyen2022revisiting}.
Single-channel images are rescaled so that their mean, plus or
minus four standard deviations, covers $[0,1]$. They are then clipped to $[0,1]$
and replicated to three channels.

\paragraph{Cost.} The table below gives the time of one evaluation on a pair of sets of $N$ images, at $32$
pixels and three channels, on a shared V100 GPU. Each entry is the median of $25$ to $30$
calls.
\begin{center}\small\setlength{\tabcolsep}{0pt}\renewcommand{\arraystretch}{1.2}
\begin{tabular*}{\columnwidth}{@{\extracolsep{\fill}}lccccc}
\toprule
 & \multicolumn{2}{c}{\colgroup{latent-based}} & \multicolumn{3}{c}{\colgroup{sliced Wasserstein}} \\
\cmidrule(lr){2-3}\cmidrule(lr){4-6}
$N$ & FID & CMMD & CSW & MSW & MSW with $p$-value \\
\midrule
200 & 1.63\,s & 4.23\,s & 0.018\,s & 0.078\,s & 0.498\,s \\
500 & 4.13\,s & 10.6\,s & 0.026\,s & 0.203\,s & 0.816\,s \\
1000 & 8.36\,s & 21.2\,s & 0.037\,s & 0.404\,s & 1.329\,s \\
\bottomrule
\end{tabular*}
\end{center}

\paragraph{Checkerboard on ImageNet.} The images are ImageNet
images of $256$ pixels. The networks of FID and CMMD resize them, which
leaves $34\,\%$ to $48\,\%$ of the amplitude of the checkerboard. As a control, we blur these images instead. CMMD then detects the change
from sets of $500$ images, so its difficulty is specific to the checkerboard.
In Figure~\ref{fig:checker256}, the power of a baseline is computed with a
Gaussian approximation of its values.

\paragraph{Power of the baselines.} The power of a baseline $D$ is
estimated in two steps. First, we compute $D(A,A_0)$ on $100$ unperturbed
runs, and $t_D$ is the $95$th percentile of these values. Second, we compute
$D(A,A_\varepsilon)$ on $12$ perturbed runs, and the power is the share of
these values that exceed $t_D$. The amplitudes $\varepsilon$ follow a
geometric grid.

For the sample requirement, we repeat the two steps at each set size $N$,
with at least $50$ unperturbed and $10$ perturbed runs. The sizes are the
powers of two from $16$ to at least $1024$.

MSW skips the first step, since its test needs no unperturbed runs. Its
power is the share of at least $24$ perturbed runs that the test rejects,
and its sizes start at $8$.

\paragraph{Comparison with the true distance.} We use $4096$ ImageNet
validation images~\cite{ref:imagenet}, reduced to $64$ pixels. In each run,
we draw two sets of $800$ images and we change the images of the second set.
The curves of Figure~\ref{fig:detail} are averages over $16$ runs. We now
prove the claim of Section~\ref{sec:detail} that removing the detail and
adding it give the same distance.
\begin{proposition}
\label{prop:detail}
Let $f:\R^d\to\R$ be convex, let
\begin{equation}
\label{eq:prox}
T(\V x)=\operatorname*{arg\,min}_{\V y}\ \tfrac12\norm{\V x-\V y}^2+f(\V y),
\end{equation}
and let $R(\V x)=2\V x-T(\V x)$. For every distribution $\mu$ on $\R^d$ with a
finite second moment, $W_2^2(\mu,T_\sharp\mu)$ and $W_2^2(\mu,R_\sharp\mu)$
are both equal to the right-hand side of~\eqref{eq:exact}.
\end{proposition}
\begin{proof}
Let $g(\V x)=\min_{\V y}\tfrac12\norm{\V x-\V y}^2+f(\V y)$. This function
is convex, as the minimum over $\V y$ of a function that is convex in
$(\V x,\V y)$. Its gradient is $\V x-T(\V x)$~\cite{ref:moreau}. The
function
\begin{equation}
\tfrac12\norm{\V x}^2-g(\V x)=\max_{\V y}\ \V x^\top\V y-\tfrac12\norm{\V y}^2-f(\V y)
\end{equation}
is also convex, as a maximum of affine functions of $\V x$. Thus, $T$ and
$R$ are the gradients of the convex functions
$\tfrac12\norm{\V x}^2-g(\V x)$ and $\tfrac12\norm{\V x}^2+g(\V x)$, and
both are optimal transport
maps~\cite[Thm.~1.48]{santambrogioOptimalTransportApplied2015}. The cost of
$T$ is the integral of $\norm{\V x-T(\V x)}^2$. The cost of $R$ is the same,
since $R(\V x)-\V x=\V x-T(\V x)$.
\end{proof}
In the left panel of Figure~\ref{fig:detail}, $f$ is a multiple of the total
variation~\cite{ref:rof}, with eight weights from $0.005$ to $0.12$. We
compute $T$ with $1000$ iterations of an accelerated projected-gradient
algorithm on the dual problem. In the right panel, $T$ is a periodic
Gaussian blur of the two colour axes of Section~\ref{sec:bank}, with seven
standard deviations from $0.75$ to $6$ pixels. It is a symmetric matrix
$\M B$ with eigenvalues in $(0,1]$, and it has the form~\eqref{eq:prox}
with $f(\V y)=\tfrac12\V y^\top(\M B^{-1}-\M I)\V y$. The
horizontal axis is the root mean square of $\V x-T(\V x)$ over the pixels of
the changed images, for images in $[0,1]$. FID, CMMD
and CSW are shown as in Figure~\ref{fig:restoration}, after the values of FID
and CSW between two sets of unchanged images are subtracted. The images with doubled
detail leave the range $[0,1]$. The networks of FID and CMMD need images in
$[0,1]$, so these images are first rescaled as described above for
single-channel images. MSW and CSW receive them as they are.

\paragraph{Generated images.} The reference set is the $30{,}000$ images of
COCO-30k (val 2014), centre-cropped and resized to $512$ pixels as in the
CMMD paper, and the generated sets are the Stable Diffusion 1.4 images for
the same captions after DDIM iterations $95$ to $100$ of $100$, decoded from
the stored latents. FID uses the clean-fid preprocessing~\cite{parmar2022aliased}, CMMD the official
configuration (CLIP ViT-L/14 at $336$ pixels), and MSW the distance
of~\eqref{eq:msw} at $512$ pixels. CSW
also uses $512$ pixels. On the full sets, the metrics read as follows.
\begin{center}\small\setlength{\tabcolsep}{0pt}
\begin{tabular*}{\columnwidth}{@{\extracolsep{\fill}}lcccccc}
\toprule
iteration & 95 & 96 & 97 & 98 & 99 & 100 \\
\midrule
FID & 15.8 & 14.1 & 13.0 & 12.5 & 13.2 & 13.4 \\
CMMD & 0.669 & 0.611 & 0.569 & 0.543 & 0.516 & 0.512 \\
MSW & 0.0537 & 0.0521 & 0.0507 & 0.0496 & 0.0485 & 0.0484 \\
\bottomrule
\end{tabular*}
\end{center}
In Figure~\ref{fig:sd}, each draw takes $N$ of the first
$4096$ captions without replacement, and the six sets
are ordered correctly when the metric decreases strictly from iteration $95$
to $100$. There are $100$ draws per size up to $N=128$, $50$ at $256$ and
$30$ at $512$. On five random subsets of the $30{,}000$ captions
per size, FID reads $494\,\%$, $300\,\%$ and $177\,\%$ of its full-set value at
$1000$, $2000$ and $5000$ images, CMMD $100\,\%$, $98\,\%$ and $100\,\%$, and MSW
$98\,\%$, $99\,\%$ and $100\,\%$.

\begin{table*}[t]\centering\footnotesize
\caption{Sample requirement $N^\ast$ of every pair of COCO classes (--: no detection).}
\label{tab:classpairs}
\setlength{\tabcolsep}{3.5pt}\renewcommand{\arraystretch}{1.12}
\resizebox{\textwidth}{!}{\begin{tabular}{lcccc@{\hspace{1.6em}}lcccc@{\hspace{1.6em}}lcccc}
\toprule
 & \multicolumn{2}{c}{\colgroup{latent-based}} & \multicolumn{2}{c}{\colgroup{sliced Wasserstein}} &  & \multicolumn{2}{c}{\colgroup{latent-based}} & \multicolumn{2}{c}{\colgroup{sliced Wasserstein}} &  & \multicolumn{2}{c}{\colgroup{latent-based}} & \multicolumn{2}{c}{\colgroup{sliced Wasserstein}} \\
\cmidrule(lr){2-3}\cmidrule(lr){4-5}\cmidrule(lr){7-8}\cmidrule(lr){9-10}\cmidrule(lr){12-13}\cmidrule(lr){14-15}
pair & FID & CMMD & CSW & MSW & pair & FID & CMMD & CSW & MSW & pair & FID & CMMD & CSW & MSW \\
\midrule
airplane--bed & 16 & 8 & 16 & 32 & boat--motorcycle & 8 & 8 & 32 & 32 & clock--truck & 16 & 8 & 128 & 32 \\
airplane--boat & 8 & 8 & 64 & 32 & boat--person & 64 & 8 & 32 & 32 & dining table--giraffe & 32 & 8 & 64 & 16 \\
airplane--bus & 8 & 8 & 32 & 16 & boat--toilet & 8 & 8 & 32 & 16 & dining table--horse & 8 & 8 & 32 & 32 \\
airplane--car & 16 & 8 & 32 & 16 & boat--train & 8 & 8 & 128 & 128 & dining table--motorcycle & 32 & 8 & 32 & 32 \\
airplane--clock & 8 & 8 & 32 & 16 & boat--truck & 16 & 8 & 128 & 64 & dining table--person & 32 & 8 & 128 & 64 \\
airplane--dining table & 32 & 8 & 8 & 16 & bus--car & 8 & 8 & 256 & 128 & dining table--toilet & 8 & 8 & 32 & 16 \\
airplane--giraffe & 8 & 8 & 16 & 16 & bus--clock & 8 & 8 & 128 & 32 & dining table--train & 32 & 8 & 32 & 32 \\
airplane--horse & 8 & 8 & 32 & 16 & bus--dining table & 32 & 8 & 16 & 32 & dining table--truck & 32 & 8 & 16 & 32 \\
airplane--motorcycle & 8 & 8 & 16 & 16 & bus--giraffe & 8 & 8 & 64 & 16 & giraffe--horse & 16 & 8 & 256 & 128 \\
airplane--person & 32 & 8 & 16 & 32 & bus--horse & 8 & 8 & 256 & 32 & giraffe--motorcycle & 8 & 8 & 128 & 32 \\
airplane--toilet & 8 & 8 & 16 & 16 & bus--motorcycle & 8 & 8 & 64 & 32 & giraffe--person & 32 & 8 & 128 & 32 \\
airplane--train & 8 & 8 & 64 & 16 & bus--person & 32 & 8 & 32 & 32 & giraffe--toilet & 8 & 8 & 64 & 16 \\
airplane--truck & 16 & 8 & 32 & 16 & bus--toilet & 8 & 8 & 32 & 16 & giraffe--train & 8 & 8 & 64 & 16 \\
bed--boat & 16 & 8 & 32 & 32 & bus--train & 8 & 8 & -- & 128 & giraffe--truck & 16 & 8 & 128 & 16 \\
bed--bus & 8 & 8 & 32 & 16 & bus--truck & 8 & 8 & -- & 256 & horse--motorcycle & 8 & 8 & 256 & 64 \\
bed--car & 8 & 8 & 64 & 32 & car--clock & 16 & 8 & 128 & 64 & horse--person & 64 & 8 & 128 & 64 \\
bed--clock & 16 & 8 & 32 & 32 & car--dining table & 8 & 8 & 32 & 32 & horse--toilet & 8 & 8 & 32 & 16 \\
bed--dining table & 32 & 8 & 128 & 32 & car--giraffe & 16 & 8 & 128 & 32 & horse--train & 8 & 8 & 256 & 64 \\
bed--giraffe & 16 & 8 & 128 & 16 & car--horse & 16 & 8 & 256 & 128 & horse--truck & 16 & 8 & -- & 64 \\
bed--horse & 16 & 8 & 64 & 16 & car--motorcycle & 32 & 8 & 128 & 64 & motorcycle--person & 64 & 8 & 128 & 32 \\
bed--motorcycle & 16 & 8 & 32 & 16 & car--person & 32 & 8 & 128 & 64 & motorcycle--toilet & 8 & 8 & 32 & 8 \\
bed--person & 32 & 8 & -- & 128 & car--toilet & 8 & 8 & 64 & 16 & motorcycle--train & 8 & 8 & 64 & 64 \\
bed--toilet & 16 & 8 & 128 & 32 & car--train & 8 & 8 & 256 & 128 & motorcycle--truck & 32 & 8 & 128 & 64 \\
bed--train & 8 & 8 & 64 & 16 & car--truck & 64 & 32 & -- & -- & person--toilet & 16 & 8 & 64 & 32 \\
bed--truck & 16 & 8 & 32 & 16 & clock--dining table & 32 & 8 & 32 & 32 & person--train & 64 & 8 & 32 & 32 \\
boat--bus & 8 & 8 & 128 & 64 & clock--giraffe & 8 & 8 & 128 & 32 & person--truck & 64 & 8 & 64 & 64 \\
boat--car & 16 & 8 & 128 & 64 & clock--horse & 8 & 8 & 128 & 64 & toilet--train & 8 & 8 & 64 & 8 \\
boat--clock & 8 & 8 & 128 & 32 & clock--motorcycle & 8 & 8 & 128 & 32 & toilet--truck & 8 & 8 & 32 & 16 \\
boat--dining table & 32 & 8 & 16 & 16 & clock--person & 32 & 8 & 128 & 128 & train--truck & 8 & 8 & -- & 256 \\
boat--giraffe & 8 & 8 & 32 & 32 & clock--toilet & 8 & 8 & 32 & 16 & & & & & \\
boat--horse & 8 & 8 & 128 & 32 & clock--train & 8 & 8 & 64 & 32 & & & & & \\
\bottomrule
\end{tabular}}
\end{table*}

\paragraph{Super-resolution.} We use $99$ images of the DIV2K validation
set~\cite{ref:div2k}, with the low-resolution images that come with it
(bicubic, factor four). From each image, we take the four
central crops of $512$ pixels, in the original and in the output. A full set
therefore has $396$ images. The weight $\alpha$ goes from $0$ to $1.4$ in
steps of $0.1$, and in steps of $0.05$ between $0.8$ and $1.2$.
MSW uses six scales, and the four crops of a scene stay in the same block.
CSW uses $2048$ slices.

In Figure~\ref{fig:restoration}(b), a draw takes two disjoint sets of $12$
scenes, which gives $48$ images per set. The originals of the first set are
compared with the outputs of the second. All metrics use the same $200$
draws.

Figure~\ref{fig:restoration} uses the two networks released with
ESRGAN~\cite{ref:esrgan}. We repeat the experiment with four other
pairs of a PSNR-trained and a GAN-trained network. They are the ESRGAN and
SRGAN~\cite{ref:srgan} pairs of the MMEditing
project~\cite{ref:mmediting}, Real-ESRGAN~\cite{ref:realesrgan} and
BSRGAN~\cite{ref:bsrgan}. The last two are trained for unknown blur, noise
and compression. The table below gives the median of three quantities
over the five pairs.
\begin{center}\small\setlength{\tabcolsep}{0pt}\renewcommand{\arraystretch}{1.2}
\begin{tabular*}{\columnwidth}{@{\extracolsep{\fill}}lccc}
\toprule
 & \colgroup{(a) same scenes} & \multicolumn{2}{c}{\colgroup{(b) different scenes}} \\
\cmidrule(lr){2-2}\cmidrule(lr){3-4}
 & minimum & minimum & draws \\
\midrule
FID & 0.74 & 0.99 & 90\,\% \\
CMMD & $-0.09$ & 0.77 & 100\,\% \\
CSW & 0.81 & 1.00 & 51.5\,\% \\
MSW & 0.12 & 0.20 & 99\,\% \\
\bottomrule
\end{tabular*}
\end{center}
The minimum is the lowest value of the curve, and closer to zero is
better. A
value of $1$ means that the curve has no minimum.
The estimate of CMMD is unbiased, so it can be slightly negative. The last
column is the share of the draws in which the metric is smaller at
$\alpha=1$ than at $\alpha=0$, and larger is better. In panel (a), the
minimum of MSW lies beyond $\alpha=1.05$ only with Real-ESRGAN and BSRGAN.

\begin{figure*}[t]
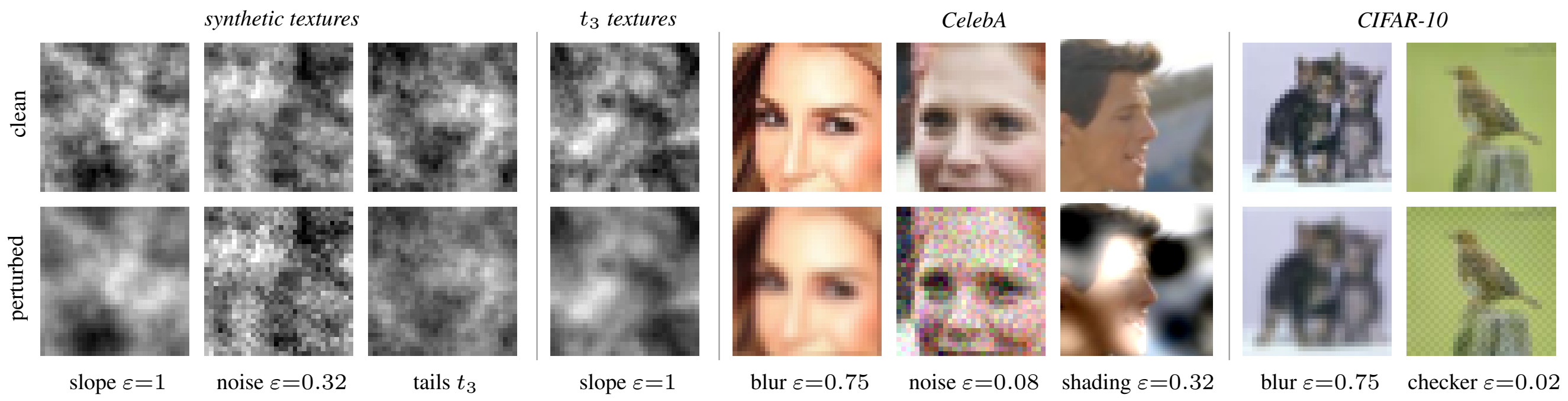

\centering
\resizebox{\textwidth}{!}{\begin{tikzpicture}[x=1mm, y=1mm, every node/.style={inner sep=0}]
\foreach \i/\x/\lab in {1/8.5/{slope $\varepsilon{=}1$}, 2/26.5/{noise $\varepsilon{=}0.32$}, 3/44.5/{tails $t_3$},
                          4/64.5/{slope $\varepsilon{=}1$},
                          5/84.5/{blur $\varepsilon{=}0.75$}, 6/102.5/{noise $\varepsilon{=}0.08$}, 7/120.5/{shading $\varepsilon{=}0.32$},
                          8/140.5/{blur $\varepsilon{=}0.75$}, 9/158.5/{checker $\varepsilon{=}0.02$}}{
    \node at (\x,-8.5)  {\includegraphics[width=17mm, height=17mm]{figures/fig1_c\i_clean.eps}};
    \node at (\x,-26.5) {\includegraphics[width=17mm, height=17mm]{figures/fig1_c\i_pert.eps}};
    \node[font=\scriptsize, anchor=base, text depth=0pt] at (\x,-38.7) {\lab};}
\node[font=\scriptsize\itshape] at (26.5,1.8) {synthetic textures};
  \node[font=\scriptsize\itshape] at (64.5,1.8) {$t_3$ textures};
  \node[font=\scriptsize\itshape] at (102.5,1.8) {CelebA};
  \node[font=\scriptsize\itshape] at (149.5,1.8) {CIFAR-10};
\foreach \x in {54.5, 74.5, 130.5} \draw[black!40, thin] (\x,-35.5) -- (\x,0.5);
\node[font=\scriptsize, rotate=90] at (-2.5,-8.5)  {clean};
  \node[font=\scriptsize, rotate=90] at (-2.5,-26.5) {perturbed};
\end{tikzpicture}
 }
\caption{Nine settings, clean and perturbed.}
\label{fig:datasets}
\end{figure*}
 \begin{table*}[t]
\centering\small
\caption{Sliced Wasserstein distance (SW) on the nine controlled perturbations (--: no detection).}
\label{tab:sw}
\begin{tabular*}{\textwidth}{@{\extracolsep{\fill}}llccc@{\hspace{3.2em}}cccc@{}}
\toprule
 & & \multicolumn{3}{c}{detection threshold $\varepsilon_{0.8}$ ($N{=}500$)} & \multicolumn{4}{c}{sample requirement $N^\ast$ at amplitude $\varepsilon$} \\
\cmidrule(r{2.4em}){3-5}\cmidrule{6-9}
 & & SW & CSW & MSW & $\varepsilon$ & SW & CSW & MSW \\
\midrule
CelebA & Gaussian noise & -- & -- & \textbf{0.0005} & 0.0025 & -- & -- & \textbf{8} \\
 & Gaussian blur & 0.75 & -- & \textbf{0.10} & 0.33 & -- & -- & \textbf{8} \\
 & shading & \textbf{0.08} & 0.16 & \textbf{0.08} & 0.08 & 512 & 1024 & \textbf{256} \\
\addlinespace
CIFAR-10 & checkerboard & 0.02 & 0.02 & \textbf{0.01} & 0.01 & 1024 & -- & \textbf{256} \\
 & Gaussian blur & -- & -- & \textbf{0.15} & 0.33 & -- & -- & \textbf{16} \\
\addlinespace
synthetic textures & spectral slope & -- & -- & \textbf{0.008} & 0.064 & -- & -- & \textbf{8} \\
 & white noise & -- & -- & \textbf{0.04} & 0.08 & -- & -- & \textbf{8} \\
 & heavier tails & 1/12 & 1/12 & \textbf{1/48} & 1/12 & 512 & 512 & \textbf{128} \\
\addlinespace
heavy-tailed textures & spectral slope & -- & -- & \textbf{0.032} & 0.128 & -- & -- & \textbf{64} \\
\bottomrule
\end{tabular*}
\end{table*}

\paragraph{Class pairs.} We use the COCO-30k photographs, centre-cropped
and resized to $512$ pixels. A photograph belongs to the class of its
largest annotated object. We keep the $14$ classes with at least $512$
photographs: person, dining table, train, bus, bed, car, toilet, truck,
clock, horse, motorcycle, giraffe, airplane and boat. Each class is limited
to $1024$ photographs, taken at random.

For a pair of classes, a run draws $N$ images of each class. An unperturbed
run draws two disjoint sets of $N$ images of the same class. At each size
$N$, there are $30$ runs per pair and $30$ unperturbed runs per class. The
power of a baseline follows the two steps above, with the unperturbed runs
of the first class of the pair. The sizes stop at $256$, since the smallest
class has $512$ photographs.

On the unperturbed runs, the test of MSW rejects $5\,\%$ of the time on
average, and at most $17\,\%$ for one class and size. The median sample
requirement over the pairs is $32$ images for MSW, $64$ for CSW, and $8$
for FID and CMMD.

\section{Details for the controlled perturbations}
\label{app:more}
Figure~\ref{fig:datasets} shows seven of the ten settings of
Table~\ref{tab:more}, and two more: a grey checkerboard on CIFAR-10 images
and the shading on CelebA images.

\begin{table*}[t]
\centering\small
\caption{Threshold $\varepsilon_{0.8}$ at $N=500$ when the features of the baselines go through the test of MSW (--: no detection).}
\label{tab:featuretest}
\begin{tabular*}{\textwidth}{@{\extracolsep{\fill}}llccc@{\hspace{2.4em}}ccc@{\hspace{2.4em}}c@{}}
\toprule
 & & \multicolumn{3}{c}{without the test} & \multicolumn{3}{c}{with the test} & \\
\cmidrule(r{1.8em}){3-5}\cmidrule(r{1.8em}){6-8}
 & & FID & CMMD & CSW & FID & CMMD & CSW & MSW \\
\midrule
CelebA & Gaussian noise & 0.01 & 0.0025 & -- & 0.005 & 0.0025 & 0.04 & \textbf{0.0005} \\
 & Gaussian blur & 0.33 & 0.22 & -- & 0.33 & 0.22 & 0.5 & \textbf{0.10} \\
\addlinespace
COCO & grey checkerboard & 0.04 & 0.04 & 0.04 & 0.02 & 0.04 & 0.04 & \textbf{0.005} \\
 & red--green checkerboard & 0.16 & 0.08 & 0.08 & 0.08 & 0.08 & 0.04 & \textbf{0.01} \\
 & shading & 0.16 & \textbf{0.04} & 0.16 & 0.08 & \textbf{0.04} & 0.16 & 0.08 \\
\addlinespace
CIFAR-10 & Gaussian blur & 0.5 & 0.22 & -- & 0.33 & 0.22 & 0.33 & \textbf{0.15} \\
\addlinespace
synthetic textures & spectral slope & 0.064 & 0.064 & -- & 0.128 & 0.064 & 0.128 & \textbf{0.008} \\
 & white noise & 0.16 & 0.08 & -- & 0.16 & 0.08 & 0.16 & \textbf{0.04} \\
 & heavier tails & 1/6 & 1/12 & 1/12 & 1/6 & 1/12 & 1/12 & \textbf{1/48} \\
\addlinespace
heavy-tailed textures & spectral slope & 0.064 & 0.064 & -- & 0.128 & 0.064 & 0.128 & \textbf{0.032} \\
\bottomrule
\end{tabular*}
\end{table*}
\paragraph{Natural images.} The CelebA images are crops of $32$ pixels.
The COCO images are $4096$ photographs of COCO-30k, centre-cropped and
resized to $512$ pixels. The blur is Gaussian, with a standard deviation of
$\varepsilon$ pixels. The checkerboard has a period of two pixels, and the
red--green pattern is the checkerboard times
$\varepsilon\,(1,-1,0)/\sqrt2$. The shading is a random smooth image with
one to three cycles per image and standard deviation $\varepsilon$, added
equally to the three channels.

For the two settings outside Table~\ref{tab:more}, FID and CMMD detect the
grey checkerboard on CIFAR-10 images from $0.01$, and the shading on CelebA
images of $64$ pixels from $0.04$. Table~\ref{tab:sw} gives the thresholds
of CSW and MSW.

\paragraph{Synthetic textures.} The textures are Gaussian random fields of
$32$ pixels, with unit variance and power spectrum
$\norm{\boldsymbol\omega}^{-3}$ at the spatial frequency
$\boldsymbol\omega$. The change of spectral slope multiplies the spectrum by
$\norm{\boldsymbol\omega}^{-\varepsilon}$ and keeps the variance. The white
noise has standard deviation $\varepsilon$. For the heavier tails, every
image is multiplied by a random scale, which gives a Student-$t$
distribution with $1/\varepsilon$ degrees of freedom.

\raggedbottom
\paragraph{Heavy-tailed textures.} The textures of the last row of
Table~\ref{tab:more} follow a Student-$t$ distribution with three degrees of
freedom. Every image is brought to zero mean and unit variance before the
metrics are applied. Without this step, the thresholds of FID and CMMD are
the same, and CSW and MSW do not detect the change.

\paragraph{Sliced Wasserstein distance.} Table~\ref{tab:sw} adds the sliced
Wasserstein distance of~\eqref{eq:sw} (SW), with $8192$ random zero-mean
directions on the vectorized images and the same estimate as CSW. SW
detects four of the nine perturbations and CSW three. Neither is ahead of
MSW in any row.

\paragraph{Ablation.} Figure~\ref{fig:ablation} uses sets of $500$ images.
In the left and middle panels, the perturbation is the Gaussian noise on
CelebA of Table~\ref{tab:more}, and the random filters are those of
Table~\ref{tab:ablation}. In the right panel, the images are Gaussian
textures of $128$ pixels with power spectrum
$\norm{\boldsymbol\omega}^{-1/2}$. The perturbation multiplies their
amplitude by $1+\varepsilon$ between one and three cycles per image, at
fixed variance.

\newpage

\section{Baselines with the test of MSW}
\label{app:featuretest}
Table~\ref{tab:featuretest} shows whether the baselines improve when their
features go through the test of MSW.

\paragraph{Set-up.} We give the test of Appendix~\ref{app:estimation} the features of a
baseline in place of the slices of MSW: the $2048$ Inception coordinates of FID, the $768$ CLIP coordinates
of CMMD, or the $2048$ projections of CSW. The settings are those of
Table~\ref{tab:more}, with $48$ perturbed runs per amplitude ($24$ on
COCO). On $100$ unperturbed runs, the test rejects between $1\,\%$ and
$8\,\%$ of the time. Without the test, FID and CMMD use the
threshold of Appendix~\ref{app:protocol} on the same runs. Four of their
values on the synthetic textures then differ from Table~\ref{tab:more} by
one step of the grid.

\paragraph{Results.} The thresholds of CMMD do not change. The threshold of
FID moves by one step of the grid, down in five rows and up in two. CSW now
detects every perturbation. MSW keeps the smallest threshold in nine rows
of ten. On the shading, CMMD is ahead with and without the test.

\end{document}